\pdfoutput=1

\documentclass[11pt]{article}

\usepackage[final]{acl}

\usepackage{times}
\usepackage{latexsym}

\usepackage[T1]{fontenc}

\usepackage[utf8]{inputenc}

\usepackage{microtype}

\usepackage{inconsolata}

\usepackage{graphicx}

\usepackage{contour}
\usepackage{booktabs}
\usepackage{multirow} 
\usepackage{amssymb}

\newcommand{\barec}{\textbf{\textsc{Barec}}}

\newcommand{\hide}[1]{}

\newcommand{\TAMAR}{{$\hbar$}}

\newcommand{\AYN}{{$\varsigma$}}

\usepackage{arabtex}

\usepackage{utf8}
\usepackage{epstopdf}

\newcommand{\BlevAlif}{{\bf 1-alif}}
\newcommand{\BlevBa}{{\bf 2-ba}}
\newcommand{\BlevJim}{{\bf 3-jim}}

\newcommand{\BlevZay}{{\bf 7-zay}}
\newcommand{\BlevHa}{{\bf 8-ha}}
\newcommand{\BlevTa}{{\bf 9-ta}}
\newcommand{\BlevYa}{{\bf 10-ya}}
\newcommand{\BlevKaf}{{\bf 11-kaf}}

\newcommand{\BlevNun}{{\bf 14-nun}}
\newcommand{\BlevSin}{{\bf 15-sin}}
\newcommand{\BlevAyn}{{\bf 16-ayn}}

\newcommand{\BlevSad}{{\bf 18-sad}}
\newcommand{\BlevQaf}{{\bf 19-qaf}}

\title{A Large and Balanced Corpus \\ for Fine-grained Arabic Readability Assessment}

\author{Khalid N. Elmadani,\textsuperscript{\textdagger} Nizar Habash,\textsuperscript{\textdagger} Hanada Taha-Thomure\textsuperscript{$\ddagger$}
  \\
  \textsuperscript{\textdagger}Computational Approaches to Modeling Language Lab, New York University Abu Dhabi\\
   \textsuperscript{$\ddagger$}Zai Arabic Language Research Centre, Zayed University\\
  \texttt{\{khalid.nabigh,nizar.habash\}@nyu.edu, Hanada.Thomure@zu.ac.ae}
  }

\begin{document}
\maketitle

\setcode{utf8}
\vocalize


\begin{abstract}

This paper introduces the Balanced Arabic Readability Evaluation Corpus (\barec),\footnote{\<بارق> \textit{bAriq} is Arabic for `very bright and glittering'.} a large-scale, fine-grained dataset for Arabic readability assessment. {\barec} consists of 69,441 sentences spanning 1+ million words, carefully curated to cover 19 readability levels, from kindergarten to postgraduate comprehension. 
The corpus balances genre diversity, topical coverage, and target audiences, offering a comprehensive resource for evaluating Arabic text complexity. The corpus was fully manually annotated by a large team of annotators. The average pairwise inter-annotator agreement, measured by Quadratic Weighted Kappa, is 81.8\%, reflecting a high level of substantial agreement.
Beyond presenting the corpus, we benchmark automatic readability assessment across different granularity levels, comparing a range of techniques. Our results highlight the challenges and opportunities in Arabic readability modeling, demonstrating competitive performance across various methods.
To support research and education, we make {\barec} openly available, along with detailed annotation guidelines and benchmark results.\footnote{\label{barec-site}\url{http://barec.camel-lab.com}}

\end{list}
\end{abstract}

\section{Introduction}

Text readability impacts understanding, retention, reading speed, and engagement \cite{DuBay:2004:principles}. Texts above a student's readability level can lead to disengagement \cite{klare1963measurement}. \newcite{Nassiri:2023} highlighted that readability and legibility depend on both external features (e.g., production, fonts) and content. Text leveling in classrooms helps match books to students' reading levels, promoting independent reading and comprehension \cite{allington2015research}. Developing readability models is crucial for improving literacy, language learning, and academic performance.

Readability levels have long been a key component of literacy teaching and learning.
One of the most widely used systems in English literacy is Fountas and Pinnell \cite{fountas2006leveled}, which employs qualitative measures to classify texts into 27 levels (A to Z+), spanning from kindergarten to adult proficiency. 
Similarly, \newcite{Taha:2017:guidelines}'s system for Arabic has 19 levels from Arabic letters \<أ> A to \<ق> Q.
%
%
These fine-grained levels are designed for pedagogical effectiveness, ensuring young readers experience gradual, measurable progress, particularly in early education (K–6) \cite{BarberKlauda2020}. A key advantage is that they can be easily mapped to coarser levels with fewer categories, which may be more efficient for broader applications in readability research and automated assessments.

In this paper we present the Balanced Arabic Readability Evaluation Corpus ({\barec}) -- a large-scale fine-grained readability assessment corpus across a broad space of genres and readability levels.
Inspired by the Taha/Arabi21 readability reference \cite{Taha:2017:guidelines}, which has been instrumental in tagging over 9,000 children's books, {\barec} seeks to establish a standardized framework for evaluating sentence-level\footnote{We use \textit{sentence} to refer to any standalone text segment, including phrases and single words (e.g., book titles).}
Arabic text readability across 19 distinct levels, ranging from kindergarten to postgraduate comprehension.

Our contributions are:
(a) \textbf{a large-scale curated corpus} with 69K+ sentences (1M+ words) spanning diverse genres; and 
(b) \textbf{benchmarking of automatic readability assessment} models across multiple granularities, including both fine-grained (19 levels) and collapsed tiered systems (e.g., five-level and three-level scales) to support various research and application needs, aligning with previous Arabic readability frameworks \cite{AlKhalil:2018:leveled,Al-Khalifa:2010:automatic}.
 


\section{Related Work}
\label{sec:related}





  

\paragraph{Automatic Readability Assessment}
Automatic readability assessment has been widely studied, resulting in numerous datasets and resources \cite{collins-thompson-callan-2004-language,pitler-nenkova-2008-revisiting,feng-etal-2010-comparison,vajjala-meurers-2012-improving,xu-etal-2015-problems,xia-etal-2016-text,nadeem-ostendorf-2018-estimating,vajjala-lucic-2018-onestopenglish,deutsch-etal-2020-linguistic,lee-etal-2021-pushing}. Early English datasets were often derived from textbooks, as their graded content naturally aligns with readability assessment \cite{vajjala-2022-trends}. However, copyright restrictions and limited digitization have driven researchers to crowdsource readability annotations from online sources \cite{vajjala-meurers-2012-improving,vajjala-lucic-2018-onestopenglish} or leverage CEFR-based L2 assessment exams \cite{xia-etal-2016-text}.

\paragraph{Arabic Readability Efforts}
Arabic readability research has focused on text leveling and assessment across various frameworks. \newcite{Taha:2017:guidelines} proposed a 19-level system for children's books based on qualitative and quantitative criteria. Other efforts applied CEFR leveling to Arabic, including the KELLY project’s frequency-based word lists, manually annotated corpora \cite{habash-palfreyman-2022-zaebuc, naous-etal-2024-readme}, and vocabulary profiling \cite{soliman2024creating}.
\citet{el-haj-etal-2024-dares} introduced DARES, a readability assessment dataset collected from Saudi school materials.
The SAMER project \cite{al-khalil-etal-2020-large} developed a lexicon with a five-level readability scale, leading to the first manually annotated Arabic parallel corpus for text simplification \cite{alhafni-etal-2024-samer}.
Automated readability assessment has also been explored through rule-based and machine learning approaches. Early models relied on surface-level features like word and sentence length \cite{al2004assessment, Al-Khalifa:2010:automatic}, while later work incorporated POS-based and morphological features \cite{Forsyth:2014:automatic, Saddiki:2018:feature}. The OSMAN metric \cite{el-haj-rayson-2016-osman} leveraged script markers and diacritization, and recent efforts \cite{liberato-etal-2024-strategies} achieved strong results using pretrained models on the SAMER corpus. 

Building on these efforts, we curated the {\barec} corpus across genres and readability levels, and manually annotated it at the sentence-level based on an adaptation of Taha/Arabi21 guidelines \cite{Taha:2017:guidelines}, offering finer-grained control and a more objective assessment of textual variation.



\begin{figure*}[t]
\centering
 \includegraphics[width=0.9\textwidth]{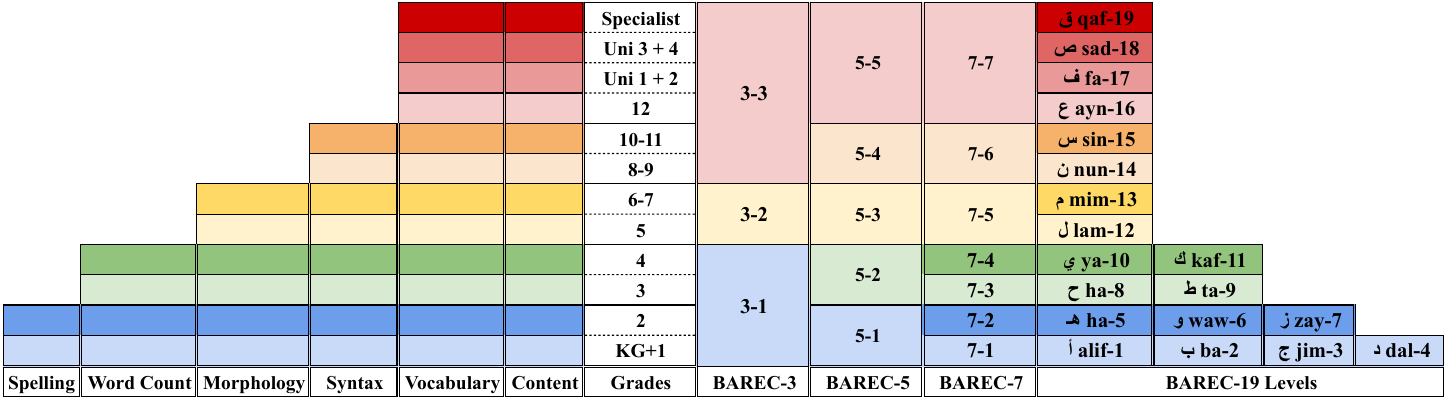}
    \caption{ The {\barec} \textit{Pyramid}  illustrates the relationship across {\barec} levels and linguistic dimensions, three collapsed variants, and education grades.}
\label{fig:barec-pyramid}
\end{figure*}

\begin{table*}[t]
\centering
 \includegraphics[width=\textwidth]{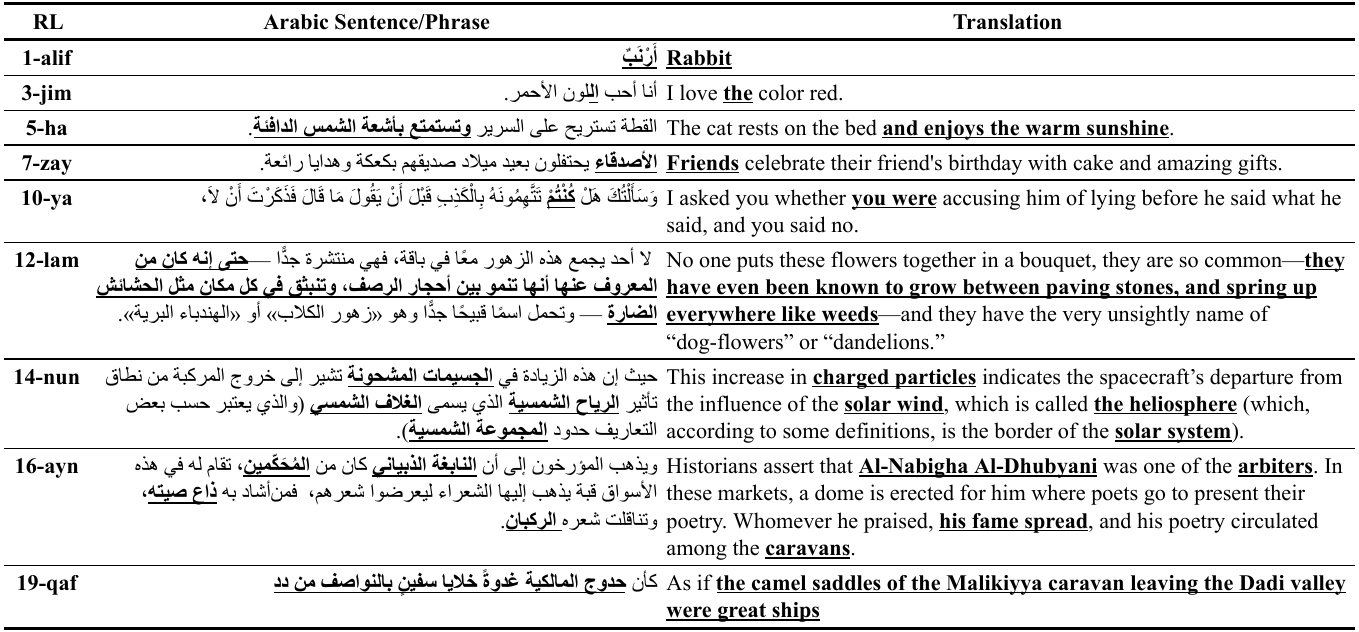}
    \caption{Representative subset of examples of the 19 {\barec} readability levels, with English translations, and readability level reasoning. Underlining is used to highlight the main keys that determined the level.  }
\label{tab:barec-examples}
\end{table*}


\section{{\barec} Corpus Annotation} \label{sec:annotation}
In this section, we summarize the guidelines and annotation process. For more details, see  \citet{habash-etal-2025-guidelines}.  In the next section, we discuss corpus selection and statistics.

\subsection{{\barec} Guidelines} 
We present below a summarized version of the {\barec} annotation guidelines. A detailed account of the  adaptation process from \newcite{Taha:2017:guidelines}'s guidelines is in \citet{habash-etal-2025-guidelines}.

\paragraph{Readability Levels}
The readability level system of \newcite{Taha:2017:guidelines} uses the Abjad order of Arabic letters for 19 levels: {\BlevAlif}, {\BlevBa}, {\BlevJim}, through to {\BlevQaf}. This system emphasizes a finer distinction in the lower levels, where readability is more varied. The {\barec} pyramid (Figure~\ref{fig:barec-pyramid}) illustrates the scaffolding of these levels and their mapping to, guidelines components, school grades, and three collapsed versions of level size 7, 5, and~3. All four level types (19-7-5-3) are fully aligned to easy mapping from fine-grained to coarse-grained levels. We present results for these levels in Section~\ref{sec:eval}. 

\paragraph{Readability Annotation Principles}
The guidelines focus on readability and comprehension, considering the ease of reading and understanding for independent readers. The evaluation does not depend on grammatical analysis or rhetorical depth but rather on understanding basic, literal meanings. Larger texts may contain sentences at different readability levels, but we focus on sentence-level evaluation, ignoring context and author intent. 

\paragraph{Textual Features}
Levels are assessed in six key dimensions.  Each of these specify numerous linguistic phenomena that are needed to qualify for being ranked in a harder level. Annotators assign each sentence a readability level based on its most difficult linguistic phenomenon. The Cheat~Sheet used by the annotators in Arabic and its translation in English are included in Appendix~\ref{app:guidelines}. 

\begin{enumerate}
    \item \textbf{Spelling}: Word length and syllable count affect difficulty.
    \item \textbf{Word Count}: The number of unique words determines the highest level for easier levels.
    \item \textbf{Morphology}: We distinguish between simple and complex morphological forms including the use of clitics and infrequent inflectional features, such as the dual.
    \item \textbf{Syntax}: Specific sentence structure and syntactic relation constructions are identified as pivotal for certain levels.
    \item \textbf{Vocabulary}: The complexity of word choices is key, with higher levels introducing more technical and classical literature vocabulary.
    \item \textbf{Content}: The required prior knowledge and abstraction levels are considered for higher levels.
\end{enumerate}
The {\barec} pyramid (Figure~\ref{fig:barec-pyramid}) illustrates which aspects are used (broadly) for which levels. For example, spelling criteria are only used up to level {\BlevZay}, while syntax is used until level {\BlevSin}, and word count is not used beyond level {\BlevKaf}.

\paragraph{Problems and Difficulties}
Annotators are encouraged to report any issues like spelling errors, colloquial language, or problematic topics. Difficulty is noted when annotations cannot be made due to conflicting guidelines.

A few representative examples for each level are provided in Table~\ref{tab:barec-examples}. A full set of examples with explanations of leveling choices is in Appendix~\ref{fullexample}. 



\begin{table*}[t]
\centering
 \includegraphics[]{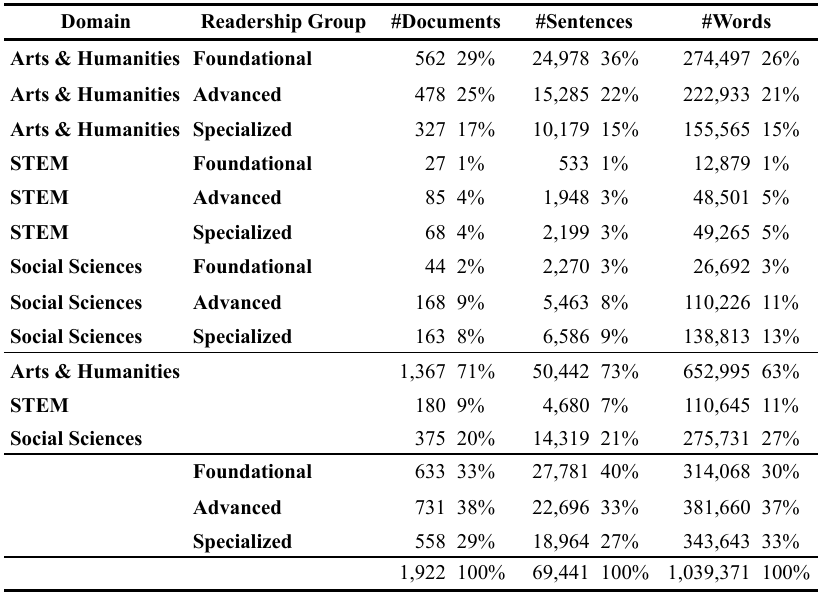}
    \caption{Summary statistics of the {\barec} Corpus.}
\label{tab:corpus-stats}
\end{table*}

\subsection{Annotation Team and Process}
 
\paragraph{Annotation Team} The {\barec} annotation team comprised six native Arabic speakers, all of whom are experienced Arabic language educators. Among the team members, one individual (A0) brought prior experience in computational linguistic annotation projects, while the remaining five (A1-5) possessed extensive expertise in readability leveling, gained through their involvement in the Taha/Arabi21 project.

\paragraph{Annotation Process} 
The annotation process began with A0, who led sentence-level segmentation and initial text flagging and selection. We followed the Arabic sentence segmentation guidelines by \newcite{habash-etal-2022-camel}.
Subsequently, A1-5 were tasked with assigning readability labels to the individually segmented texts.
The annotation  was done through a simple Google Sheet interface. 
A1-5 received folders containing  annotation sets, comprising 100 randomly selected sentences each. 
The average annotation speed was around 2.5 hours per batch (1.5 minutes/sentence).

Before starting the annotation, all annotators received rigorous training, including three pilot rounds.  These rounds provided opportunities for detailed discussions of the guidelines, helping to identify and address any issues. 
19 shared annotation sets (100 sentence each) were included covertly to ensure quality and measure inter-annotator agreement (IAA).
Finally, we conducted a thorough second review of the corpus data, resulting in every sentence being checked twice for the first phase (10,658 sentences) before continuing to finish the 69,441 sentences (1M words).

In total, the annotators annotated 92.6K sentences, 25\%  of which is not in the final corpus: 3.3\% were deemed problematic (typos and offensive topics); 11.5\% were part of the second round of first phase annotation; and 10.3\% were part of the IAA efforts, not including their unification. 
%
We report on IAA in Section~\ref{sec:iaa-results}.

\section{{\barec} Corpus}
\paragraph{Corpus Selection}

In the process of corpus selection, we aimed to cover a wide educational span as well as different domains and topics. We collected the corpus from 1,922 documents, which we manually categorized into three domains: \textbf{Arts \& Humanities}, \textbf{Social Sciences}, and \textbf{STEM} (details in Appendix~\ref{app:domains}) and three readership groups: \textbf{Foundational}, \textbf{Advanced}, and \textbf{Specialized}  (details in  Appendix~\ref{app:reader}).
Table~\ref{tab:corpus-stats} shows the distribution of the documents, sentences, and words across domains and groups.  
The distribution across readership levels aligns with the corpus's educational focus, with a higher-than-usual proportion at foundational levels. Variations across domains reflect differences in the availability of texts and reader interest (more Arts \& Humanities, less STEM).
The corpus uses documents from 30 different resources. All selected texts are either out of copyright, within the fair-use limit, or obtained in agreement with publishers. The decision of selecting some of these resources is influenced by the fact that other annotations exist for them. Around 25\% of all sentences came from completely new sources that were manually typed to make them digitally usable. 
%
All details about the resources are available in Appendix~\ref{app:full-data}.

\begin{figure*}[t]
\centering
 \includegraphics[width=1.3\columnwidth]{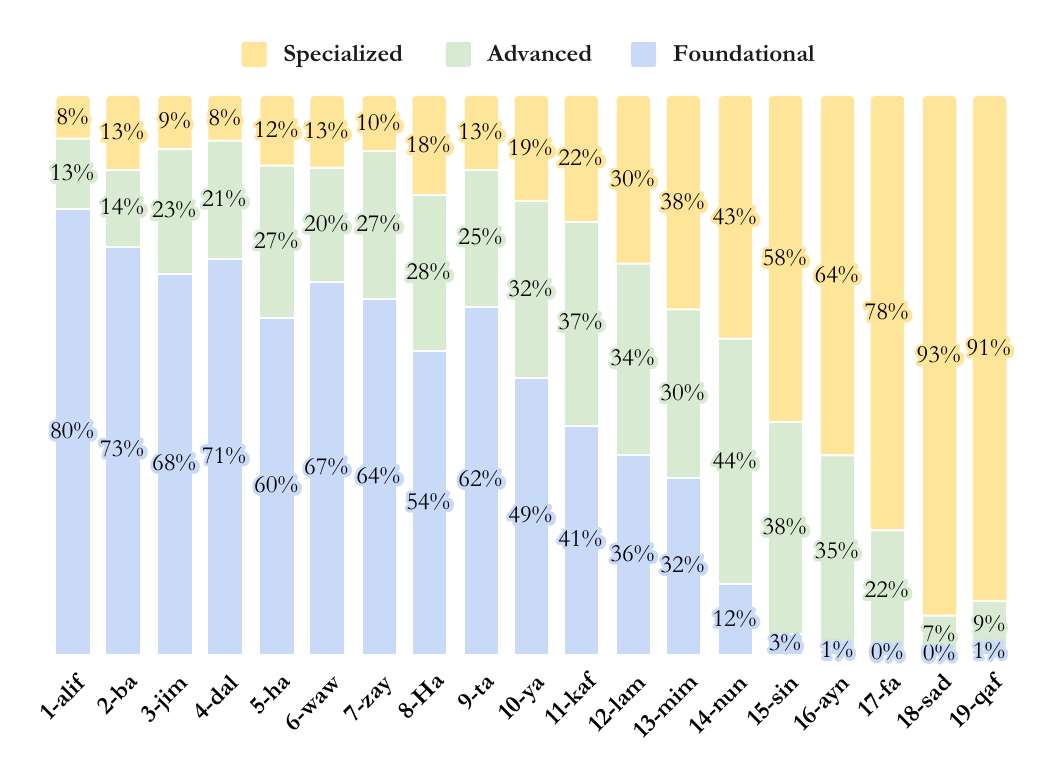}
    \caption{The distribution of the readership groups across {\barec} levels.}
\label{tab:readability-stats}
\end{figure*}

\begin{table}[t]
\centering
 \includegraphics[width=\columnwidth]{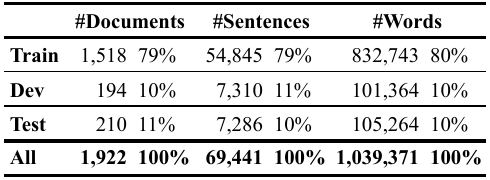}
    \caption{{\barec} Corpus splits.}
\label{tab:corpus-splits}
\end{table}

\paragraph{Readability Statistics}
Figure \ref{tab:readability-stats} shows the distribution of the three readership groups across all readability levels. As expected, foundational texts strictly dominate the lower levels up to {\BlevTa}, then the presence of advanced and specialized texts starts increasing gradually till the highest level. Specialized texts dominate the highest levels, while the middle levels ({\BlevYa} to {\BlevNun}) include a mix of the three groups with a slight advantage for advanced texts.

\paragraph{Corpus Splits}
We split the corpus into \textbf{Train ($\simeq$80\%)}, \textbf{Dev ($\simeq$10\%)}, and \textbf{Test ($\simeq$10\%)} at the document level. 
Sentences from IAA studies are divided between all splits.
However, We will release the IAA studies as a special set as they provide multiple references from different annotators for each example.\textsuperscript{\ref{barec-site}} 
Also, if other annotations exist for a resource (e.g., CamelTB \citep{CamelTB} and ReadMe++ \citep{naous-etal-2024-readme}), we follow the existing splits. Table \ref{tab:corpus-splits} shows the corpus splits in the level of documents, sentences, and words. 
More details about the splits across readability levels, domains, and readership groups are available in Appendix \ref{app:splits}.

\section{Experiments}

\subsection{Metrics}
In this paper, we define the task of Readability Assesment as an ordinal classification task. We use the following metrics for evaluation.

\paragraph{\textbf{Accuracy (Acc$^{19}$)}} 
The percentage of cases where reference and prediction classes match in the 19-level scheme. We addition consider three variants, \textbf{Acc$^{7}$}, \textbf{Acc$^{5}$}, \textbf{Acc$^{3}$}, that respectively collapse the 19-levels into the 7, 5, and 3-level schemes discussed in Section~\ref{sec:annotation}.



\paragraph{\textbf{Adjacent Accuracy ($\pm$1 Acc$^{19}$)}}
Also known as off-by-1 accuracy. It allows some tolerance for predictions that are close to the true labels. It measures the proportion of predictions that are either exactly correct or off by at most one level.

\paragraph{\textbf{Average Distance (Dist)}}
Also known as Mean Absolute Error (MAE), it measures the average absolute difference between predicted and true labels.

\paragraph{\textbf{Quadratic Weighted Kappa (QWK)}}
An extension of Cohen’s Kappa \citep{Cohen:1968:weighted,2023.EDM-long-papers.9} measuring the agreement between predicted and true labels, but applies a quadratic penalty to larger misclassifications, meaning that predictions farther from the true label are penalized more heavily.

We consider Quadratic Weighted Kappa as the primary metric for selecting the best system.

\begin{table}[t]
\centering
 \includegraphics[width=\columnwidth]{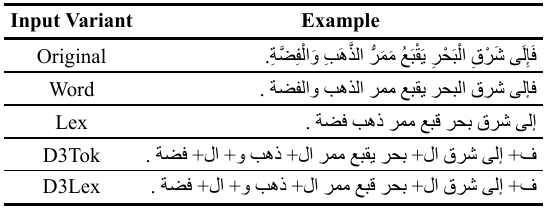}
    \caption{Example sentence in different input variants.}
\label{tab:variants-example}
\end{table}

\subsection{Input Variants}
In morphologically rich languages, affixation, compounding, and inflection convey key linguistic information that influences readability. Human annotators consider morphological complexity when assessing readability, but standard tokenization may obscure these cues. Segmenting sentences into morphological units helps preserve structural patterns relevant to readability prediction.

\hide{
To generate all input variants, we start by running all sentences through the state-of-the-art MSA morphological disambiguator \citep{inoue-etal-2022-morphosyntactic}, then selecting that top1 set of tags for all words. We use the predicted morphological features to craft the following input variants:

\paragraph{\textbf{Word}}
We tokenize the sentences and remove diacritics and kashida using CAMeL Tools \citep{obeid-etal-2020-camel}.

\paragraph{\textbf{Lex}}
We replace each word with its predicted Lemma.

\paragraph{\textbf{D3Tok}}
We replace each word with its detokenized form.

\paragraph{\textbf{D3Lex}}
Same as D3tok but the basic stem of each word is replaced with the Lemma.
\\

Table \ref{tab:variants-example} shows an example of a sentence and the corresponding input variants.
}

We generate four input variants using CamelTools  morphological disambiguation to identify top choice analysis in context \citep{obeid-etal-2020-camel}.\footnote{CamelTools v1.5.5: Bert-Disambig\textbf{+}calima-msa-s31 db.}  For the \textbf{Word} variant, we simply tokenize the sentences and remove diacritics and kashida using CAMeL Tools \citep{obeid-etal-2020-camel}.  For \textbf{Lex}, we replace each word with its predicted Lemma. For \textbf{D3Tok}, we tokenize the word into its base and clitics form; and for \textbf{D3Lex}, we replace the base form in D3Tok with the lemma. All variants are dediacritized. 
Table \ref{tab:variants-example} shows an example of a sentence and the corresponding input variants.

\subsection{Fine-Tuning}
We fine-tuned the top three Arabic BERT-based models according to \citet{inoue-etal-2021-interplay} (AraBERTv02 \citep{antoun-etal-2020-arabert}, MARBERTv2 \citep{abdul-mageed-etal-2021-arbert}, CamelBERT-msa \citep{inoue-etal-2021-interplay}). We also added AraBERTv2 to our experiments due to the possible matching between its pre-training data (morphologically segmented sentences by Farasa \citep{darwish-mubarak-2016-farasa}) and the different input variants.

\subsection{Loss Functions}
Since readability levels exhibit a natural ordering, we explore loss functions that account for the distance between predicted and true labels \citep{heilman-etal-2008-analysis}. In addition to standard cross-entropy loss (\textbf{CE}), we experiment with Ordinal Log Loss (\textbf{OLL}) \citep{castagnos-etal-2022-simple}, Soft Labels Loss (\textbf{SOFT}) \citep{9156542}, Earth Mover’s Distance-based loss (\textbf{EMD}) \citep{HouNIPSW17}, and Regression using Mean Squared Error (\textbf{Reg}) as these have been previously used for ordinal classification tasks. OLL, SOFT, and EMD incorporate a distance matrix $D$ into their formulations to penalize predictions proportionally to their distance from the true label. For simplicity, we define the distance between any two adjacent levels as one, setting $D(i,j) = |i-j|$ for labels i and j. For regression, we round the final output to the nearest readability level to ensure predictions align with the 19 levels.

\subsection{Hyper-parameters}
For all experiments, we use a learning rate of \(5 \times 10^{-5}\), a batch size of $64$, and train for six epochs on an NVIDIA V100 GPU. After training, we select the best-performing epoch based on evaluation loss. For Ordinal Log Loss (OLL), we experiment with different values of the weighting parameter \(\alpha\), choosing from \{0.5, 1, 1.5, 2\}. Similarly, for Soft Labels Loss (SOFT), we evaluate different values of the smoothing parameter \(\beta\), selecting from \{2, 3, 4, 5\}. The training of the models in this paper took approximately 20 hours.

\subsection{Procedure}
Our experiments involve three main variables: the pretrained model, the input variant, and the loss function. Our goal is to determine the optimal combination of these three factors. Due to the large number of experiments required, we divide the process into two stages.  
In \textbf{Stage 1}, we train all combinations of pretrained models and input variants using cross-entropy loss. We then select the best combination based on a majority vote from our primary evaluation metrics (Acc, Acc~$\pm$1, Dist, and QWK).  
In \textbf{Stage 2}, we take the best combination of pretrained model and input variant from the first stage and train models using all the different loss functions. 


\section{Results}
\label{sec:eval}
\subsection{Inter-Annotator Agreement (IAA)}
\label{sec:iaa-results}

In this section, we report on  16   IAA studies, excluding the three pilots and first two IAAs, which overlapped with annotator training.

\paragraph{Pairwise Agreement}
The average pairwise exact-match over 19 {\barec} levels between any two annotators is only 61.1\%, which reflects the task's complexity. Allowing a fuzzy match distance of up to one level raises the match to 74.4\%.
The overall average pairwise level difference is 0.94 levels.
The average pairwise Quadratic Weighted Kappa 81.8\% 
(substantial agreement) confirms most disagreements are minor \cite{Cohen:1968:weighted,2023.EDM-long-papers.9}.  

\paragraph{Unification Agreement}
After each IAA study, the annotators discussed and agreed on a unified readability level for each sentence. On average, the exact match between the annotators and the unified level (Acc$^{19}$) was 71.7\%, reflecting the difficulty of the task. However, the high average $\pm$1 Acc$^{19}$ (82.3\%), low Distance (0.65), and strong Quadratic Weighted Kappa (88.1\%) suggest that most disagreements between annotators and the unified labels were minor.
For more detailed results on IAA, see \cite{habash-etal-2025-guidelines}.

\begin{table*}[t]
\centering
\begin{tabular}{llcccc}
\toprule
\textbf{Input} & \textbf{Model} & \textbf{Acc$^{19}$} & \textbf{$\pm$1 Acc$^{19}$} & \textbf{Dist} & \textbf{QWK} \\
\midrule
\multirow{4}{*}{Word}
  & CamelBERT-msa & 54.4\% & 68.7\% & 1.20 & 79.1\% \\
  & MARBERTv2     & 53.3\% & 68.0\% & 1.20 & 79.1\% \\
  & AraBERTv02    & 55.8\% & 69.2\% & 1.17 & 79.2\% \\
  & AraBERTv2     & 51.6\% & 65.9\% & 1.32 & 76.3\% \\
\midrule
\multirow{4}{*}{Lex}
  & CamelBERT-msa & 48.3\% & 64.4\% & 1.34 & 77.1\% \\
  & MARBERTv2     & 50.1\% & 64.9\% & 1.31 & 77.0\% \\
  & AraBERTv02    & 48.8\% & 65.4\% & 1.30 & 78.5\% \\
  & AraBERTv2     & 50.1\% & 65.4\% & 1.29 & 77.7\% \\
\midrule
\multirow{4}{*}{D3Tok}
  & CamelBERT-msa & 54.8\% & 68.2\% & 1.21 & 78.2\% \\
  & MARBERTv2     & 54.0\% & 68.5\% & 1.20 & 78.9\% \\
  & AraBERTv02    & 54.8\% & 68.1\% & 1.22 & 78.2\% \\
  & AraBERTv2     & \textbf{56.6\%} & \textbf{69.9\%} & \textbf{1.14} & \textbf{80.0\%} \\
\midrule
\multirow{4}{*}{D3Lex}
  & CamelBERT-msa & 51.1\% & 65.5\% & 1.29 & 78.0\% \\
  & MARBERTv2     & 51.6\% & 65.7\% & 1.28 & 78.0\% \\
  & AraBERTv02    & 53.3\% & 68.1\% & 1.24 & 78.2\% \\
  & AraBERTv2     & 53.2\% & 67.1\% & 1.24 & 78.6\% \\
\bottomrule
\end{tabular}
\caption{Results comparing different combinations of models and input variants on {\barec} Dev set. \textbf{Bold} are the best results on each metric.}
\label{tab:model-selection}
\end{table*}

\begin{table}[t]
\centering
\begin{tabular}{lcccc}
\toprule
\textbf{Loss} & \textbf{Acc$^{19}$} & \textbf{$\pm$1 Acc$^{19}$} & \textbf{Dist} & \textbf{QWK} \\
\midrule
        CE            & \textbf{56.6\%} & 69.9\% & 1.14 & 80.0\% \\
        EMD           & 55.3\% & 70.3\% & \textbf{1.11} & 81.2\% \\
        OLL2          & 35.2\% & 70.3\% & 1.25 & 82.0\% \\
        OLL15         & 47.3\% & 71.1\% & 1.13 & 82.8\% \\
        OLL1          & 50.8\% & 71.5\% & 1.12 & 81.7\% \\
        OLL05         & 53.1\% & 68.8\% & 1.18 & 79.7\% \\
        SOFT2         & 55.8\% & 69.8\% & 1.15 & 80.0\% \\
        SOFT3         & 56.4\% & 69.9\% & 1.14 & 80.1\% \\
        SOFT4         & 56.4\% & 69.9\% & 1.15 & 79.6\% \\
        SOFT5         & 56.2\% & 69.5\% & 1.17 & 79.3\% \\
        Reg     & 43.1\% & \textbf{73.1\%} & 1.13 & \textbf{84.0\%} \\
\bottomrule
\end{tabular}
\caption{Loss functions comparisons on {\barec} Dev set. We use AraBERTv2 model and D3Tok input with all loss function. \textbf{Bold} are the best results on each metric.}
\label{tab:loss-functions}
\end{table}


\begin{table*}[t]
\centering
\begin{tabular}{llcccccccc}
\toprule
\textbf{Input} & \textbf{Loss} & \textbf{Acc$^{19}$} & \textbf{$\pm$1 Acc$^{19}$} & \textbf{Dist} & \textbf{QWK} & \textbf{Acc$^{7}$} & \textbf{Acc$^{5}$} & \textbf{Acc$^{3}$} \\
\midrule
\textbf{Word}       & \textbf{CE}         & 51.6\% & 65.9\% & 1.32 & 76.3\% & 61.6\% & 67.2\% & 74.0\% \\
\textbf{Lex}        & \textbf{CE}         & 50.1\% & 65.4\% & 1.29 & 77.7\% & 60.6\% & 66.3\% & 74.9\% \\
\textbf{D3Tok}      & \textbf{CE}         & \textbf{56.6\%} & 69.9\% & 1.14 & 80.0\% & 65.9\% & 70.3\% & 76.5\% \\
\textbf{D3Lex}      & \textbf{CE}         & 53.2\% & 67.1\% & 1.24 & 78.6\% & 63.6\% & 69.0\% & 75.3\% \\
\midrule
\textbf{D3Tok}      & \textbf{EMD}       & 55.3\% & 70.3\% & \textbf{1.11} & 81.2\% & 65.2\% & 70.0\% & 76.4\% \\
\textbf{D3Tok}      & \textbf{Reg} & 43.1\% & \textbf{73.1\%} & 1.13 & \textbf{84.0\%} & 61.1\% & 67.8\% & 75.9\% \\
\midrule
\multicolumn{2}{l}{\textbf{Average}}       & 46.9\% & 72.5\% & \textbf{1.11} & 83.4\% & 64.0\% & 70.3\% & \textbf{77.2\%} \\
\multicolumn{2}{l}{\textbf{Most Common}}   & 56.3\% & 70.0\% & 1.13 & 80.4\% & \textbf{66.3\%} & \textbf{70.9\%} & 76.9\% \\
\midrule
\multicolumn{2}{l}{\textbf{Oracle Combo}}  & 75.2\% & 87.4\% & 0.50 & 93.8\% & 83.2\% & 85.7\% & 89.1\% \\
\bottomrule
\end{tabular}
\caption{Results comparing different loss function, ensemble methods, and oracle performance on {\barec} Dev set. \textbf{Bold} are the best results across individual models and across ensembles.}
\label{tab:dev-results}
\end{table*}

\begin{table*}[t]
\centering
\begin{tabular}{llcccccccc}
\toprule
\textbf{Input} & \textbf{Loss} & \textbf{Acc$^{19}$} & \textbf{$\pm$1 Acc$^{19}$} & \textbf{Dist} & \textbf{QWK} & \textbf{Acc$^{7}$} & \textbf{Acc$^{5}$} & \textbf{Acc$^{3}$} \\
\midrule
\textbf{Word}       & \textbf{CE}         & 51.1\% & 65.1\% & 1.31 & 76.2\% & 60.7\% & 65.6\% & 72.2\% \\
\textbf{Lex}        & \textbf{CE}         & 51.2\% & 66.2\% & 1.23 & 78.5\% & 61.1\% & 66.2\% & 74.4\% \\
\textbf{D3Tok}      & \textbf{CE}         & 55.9\% & 70.0\% & 1.12 & 80.2\% & 65.1\% & 69.4\% & 75.2\% \\
\textbf{D3Lex}      & \textbf{CE}         & 53.7\% & 67.9\% & 1.17 & 79.5\% & 63.8\% & 69.1\% & 74.8\% \\
\midrule
\textbf{D3Tok}      & \textbf{EMD}       & 54.9\% & 71.4\% & \textbf{1.02} & 83.7\% & 64.9\% & 69.0\% & 75.2\% \\
\textbf{D3Tok}      & \textbf{Reg} & 41.4\% & \textbf{73.5\%} & 1.11 & 84.4\% & 59.4\% & 65.3\% & 72.8\% \\
\midrule
\multicolumn{2}{l}{\textbf{Average}}       & 46.0\% & 73.4\% & 1.06 & \textbf{84.5\%} & 63.6\% & 69.4\% & \textbf{75.8\%} \\
\multicolumn{2}{l}{\textbf{Most Common}}   & \textbf{56.2\%} & 70.4\% & 1.07 & 81.3\% & \textbf{65.9\%} & \textbf{70.0\%} & 75.6\% \\
\midrule
\multicolumn{2}{l}{\textbf{Oracle Combo}}  & 75.9\% & 87.8\% & 0.46 & 94.7\% & 83.5\% & 85.7\% & 88.9\% \\
\bottomrule
\end{tabular}
\caption{Results comparing different loss function, ensemble methods, and oracle performance on {\barec} Test set. \textbf{Bold} are the best results across individual models and across ensembles.}
\label{tab:test-results}
\end{table*}


\subsection{Stage 1 Results}  

Table~\ref{tab:model-selection} presents the results of stage 1, where we evaluate different combinations of pretrained models and input variants using cross-entropy loss. Based on the all metrics, we observe that the AraBERTv02 and AraBERTv2 models generally achieve higher performance across multiple input variants.  

Among input variants, the Word and D3Tok representations tend to yield better results compared to Lex and D3Lex. Specifically, AraBERTv2 with the D3Tok input achieves the best scores in all metrics.
Notably, AraBERTv2 is the only model that benefits from the D3Tok and D3Lex inputs compared to the Word input, showing an improvement across all metrics. We argue that this occurs because AraBERTv2 is the only model in this set that was pretrained on segmented data, making it more compatible with morphologically segmented input.
These results suggest that both the choice of input variant and the pretrained model significantly impact performance.

Based on all metrics, we select AraBERTv2 with the D3Tok input as the best-performing combination. In stage 2, we evaluate it with different loss functions.
The confusion matrix for this model is available in the Appendix \ref{app:conf-matrix}.

\hide{

\begin{table}[t]
\centering
 \includegraphics[width=\columnwidth]{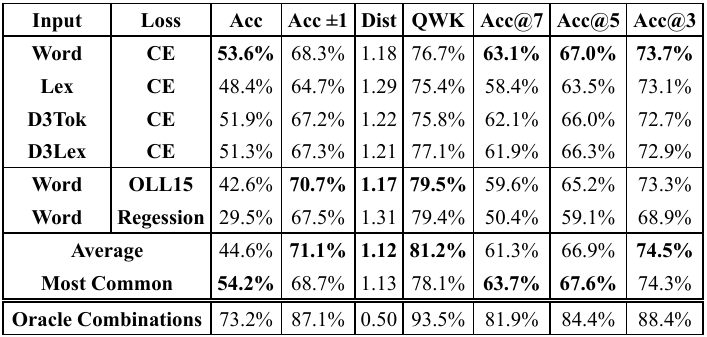}
    \caption{Dev Results}
\label{tab:dev-results}
\end{table}

\begin{table}[t]
\centering
 \includegraphics[width=\columnwidth]{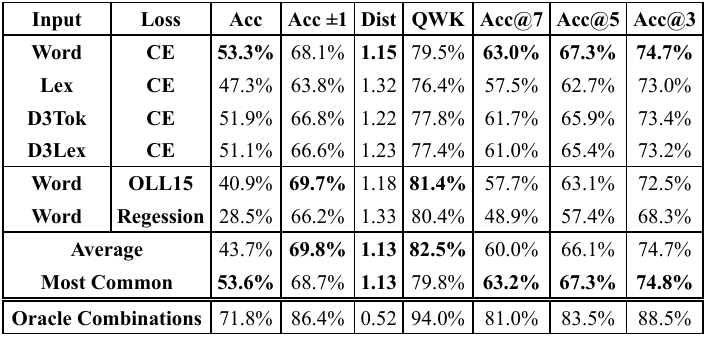}
    \caption{Test Results}
\label{tab:test-results}
\end{table}
}

\subsection{Stage 2 Results}
Table \ref{tab:loss-functions} presents the results of stage 2, where we use the best model from stage 1 to evaluate different loss functions.
Among all the loss functions evaluated, Cross-Entropy (CE) achieves the highest exact accuracy (Acc$^{19}$) at 56.6\%, indicating that it performs best when predicting the exact readability level. In contrast, other loss functions show stronger performance on metrics that consider the ordinal nature of readability levels. Notably, Regression achieves the highest $\pm$1 accuracy at 73.1\% and the best Quadratic Weighted Kappa (QWK) at 84.0\%, suggesting it excels at predicting levels close to the gold label, despite being the worst in terms of exact accuracy. These findings support that loss functions designed for ordinal or continuous labels—such as EMD, OLL, and Regression—are more effective on evaluation metrics that reward proximity to the correct label, even if they underperform on strict accuracy.
More results for other loss functions are in Appendix \ref{app:loss}.

\subsection{Ensemble Results} 
Table~\ref{tab:dev-results} presents results from Stage 1, where AraBERTv2 is evaluated with four different input variants, and Stage 2, where it is trained using the two best-performing loss functions. It also includes results from two ensemble strategies applied across all six models to assess whether combining predictions can further improve performance.
We also include an oracle combination, which represents an upper bound on performance. This allows us to estimate the maximum potential gain achievable through ensembling.


\paragraph{Ensemble}
To further improve performance, we experiment with ensemble methods. We define the \textbf{Average ensemble}, where the final prediction is the rounded average of the levels predicted by the six models, and the \textbf{Most Common ensemble}, where the final prediction is the predicted levels' mode.

The results show that the Average ensemble performs better in terms of Distance, indicating that it tends to stay closer to the correct label. However, it struggles with exact accuracy (Acc), as averaging can blur distinctions between classes.
On the other hand, the Most Common ensemble achieves higher Acc but can sometimes be misled by an incorrect majority, leading to greater deviation from the correct label.

\paragraph{Oracle}
We also report an \textbf{Oracle Combination}, where we assume access to the best possible prediction from the six models for each sample. This serves as an upper bound on model performance. The Oracle results are significantly higher than those of individual models and are comparable to human annotators’ agreement with the unified labels (see section \ref{sec:iaa-results}). This suggests that while individual models are still far from human-level performance, ensembling has the potential to push results closer to human agreement. More oracle combinations are provided in Appendix \ref{app:ensemble}.
We also include more results on the impact of training granularity on readability level prediction in Appendix \ref{app:gran}

Finally, table \ref{tab:test-results} shows the results on the test set. We note that the trends observed in the development set persist in the test set, further validating our findings. 

\hide{

\section{Results}
\label{sec:eval}

\begin{figure*}[ht!]
\centering
 \includegraphics[width=0.85\textwidth]{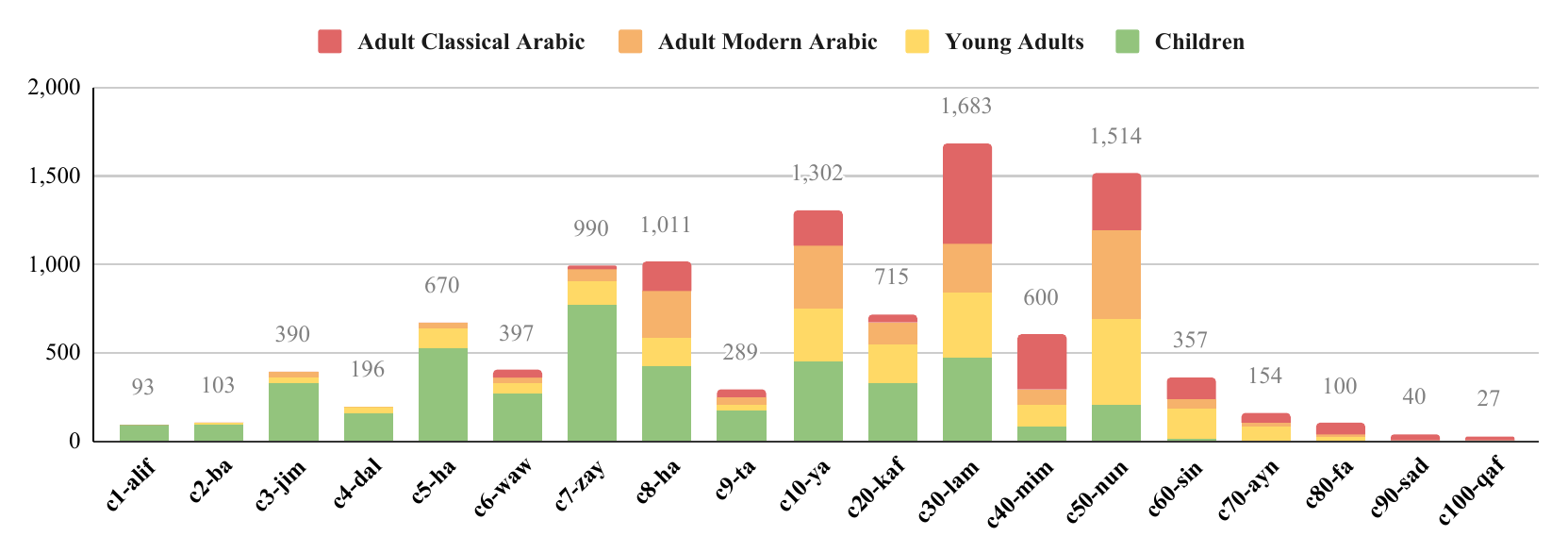}
    \caption{The distribution of annotated sentences among {\barec} levels and Arabic readers groups}
\label{fig:annotation-distribution}
\end{figure*}

\begin{figure}[ht!]
\centering
 \includegraphics[width=\columnwidth]{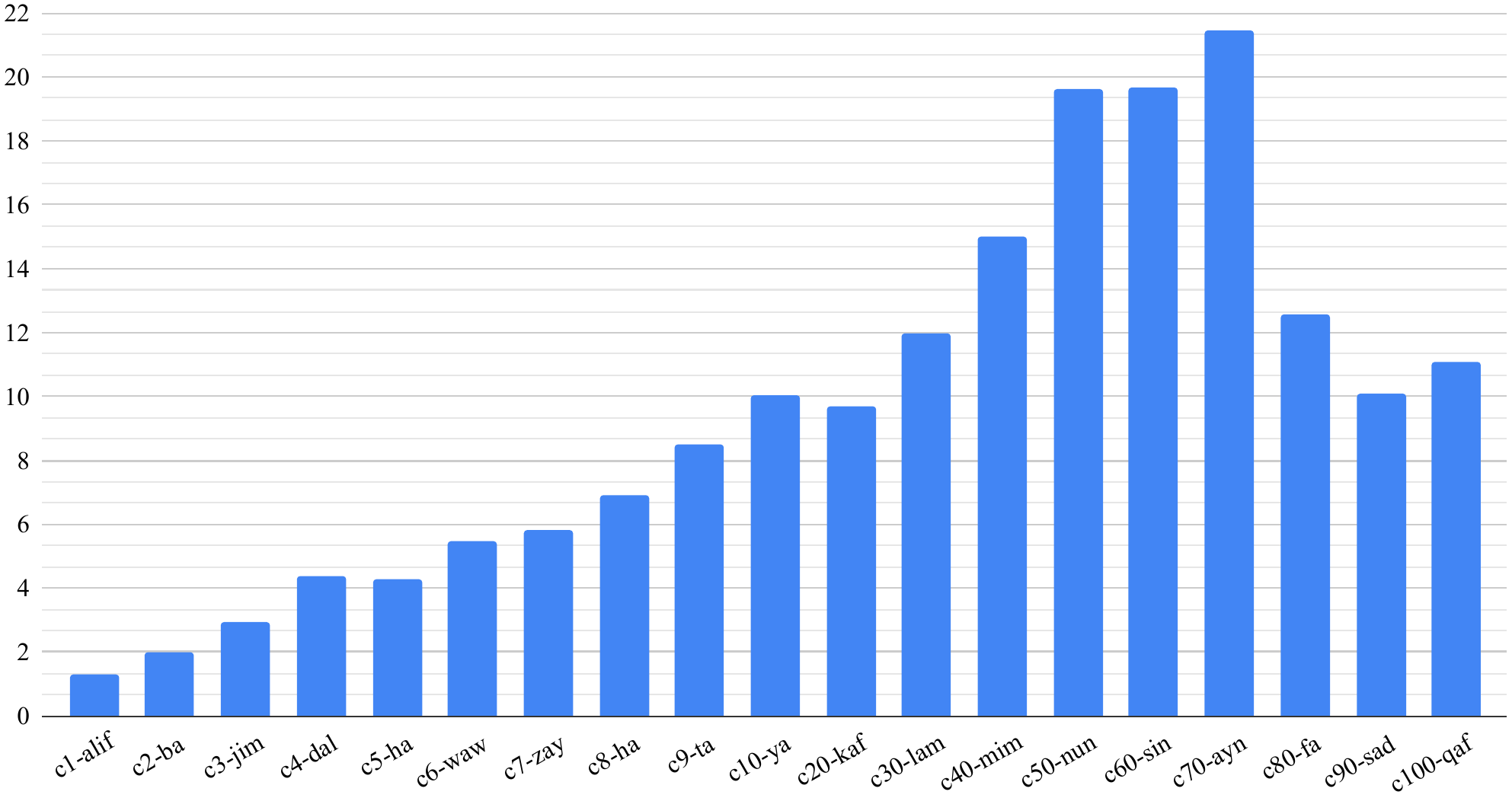}\\
  \includegraphics[width=\columnwidth]{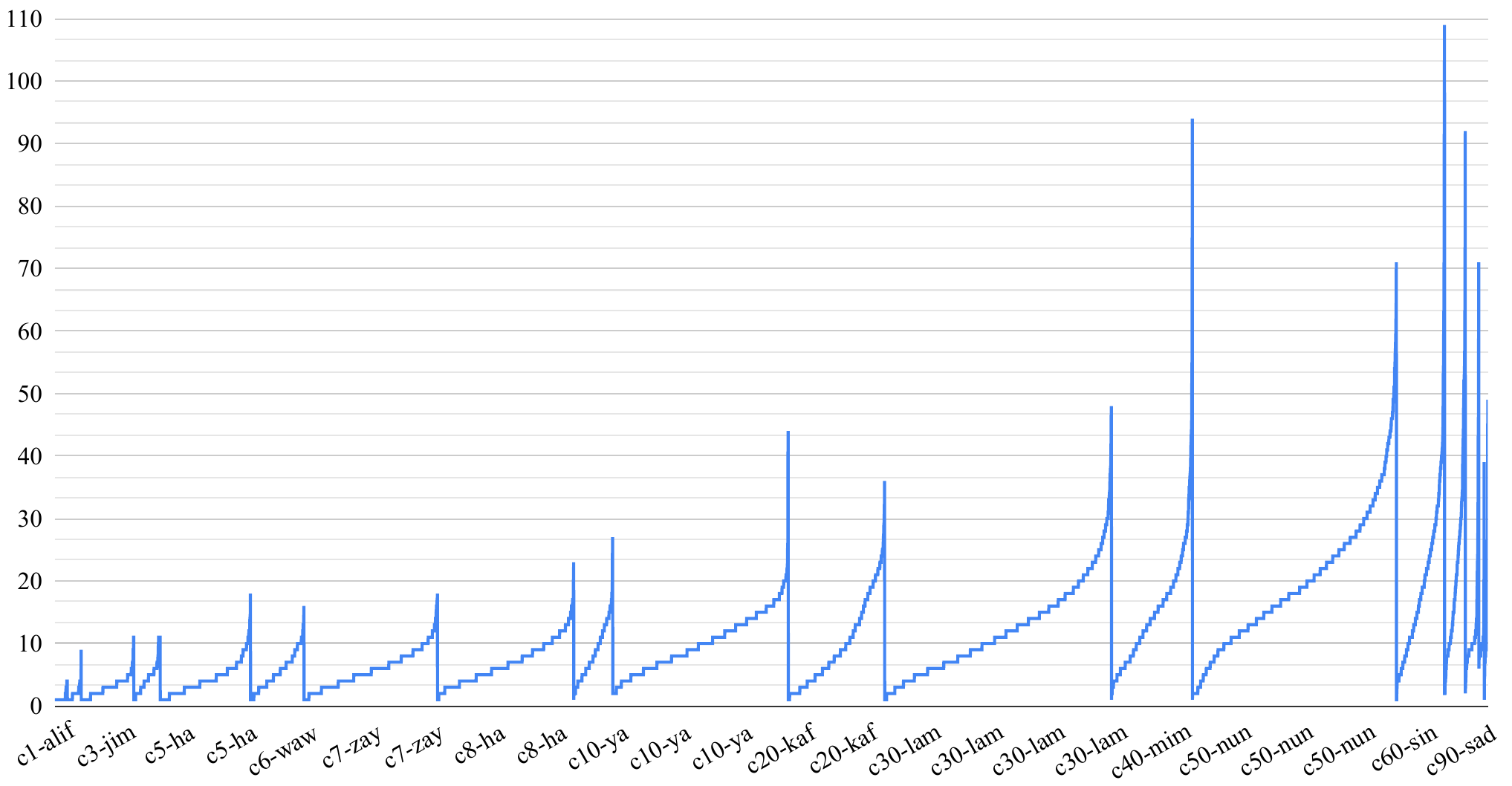}
    \caption{Charts comparing the average sentence length (left) and the distribution of lengths (right) per level}
\label{fig:annotation-level-avg}
\end{figure}

\subsection{Inter-Annotator Agreement}
%
We conducted four inter-annotator agreement (IAA) studies: three 100-sentence pilots during \textit{training} to enhance agreement, and a final official study using $200$ sentences, which we report on next.
The average pairwise exact-match over 19 {\barec} levels between any two annotators is only 49.2\%, which reflects the task's complexity. Allowing a fuzzy match distance of up to 1, 2, 3, or 4 levels raises the match to 64.6\%, 77.1\%, 87.2\%, and 93.2\%, respectively. 
The overall average pairwise level difference is $1.38$ levels. 
The average pairwise Quadratic Weighted Kappa 79.9\% 
(substantial agreement) confirms most disagreements are minor \cite{Cohen:1968:weighted,2023.EDM-long-papers.9}.
%

%

\paragraph{Second Round QC}
After the above-mentioned IAA, we made some minor guideline clarifications and did some continued training.
Then we conducted a \textbf{second round of full annotation quality check} where every example was checked by a different annotator from the first round. In total  40\% of the labels changed with an average level distance of 0.97; the average pairwise Quadratic Weighted Kappa between the two rounds is 85.5\%.

\begin{table}
\centering
\setlength{\tabcolsep}{2pt}
\begin{small}

\begin{tabular}{|l|r|r|r|l|}
\hline
\textbf{Group} & \textbf{\#Docs} & \textbf{\#Sents} & \textbf{\#Words} \\ \hline
\textbf{Children} & 30 & 4,363 & 30,502 \\ \hline
\textbf{Young Adults }& 42 & 2,307 & 29,465 \\ \hline
\textbf{Adult Modern Arabic} & 74 & 1,952 & 26,108 \\ \hline
\textbf{Adult Classical Arabic} & 128 & 2,009 & 27,576 \\ \hline
\textbf{Total} & \textbf{274} & \textbf{10,631} & \textbf{113,651} \\  
\hline
\end{tabular}
\end{small}
\caption{Summary statistics of the {\barec} Corpus}
\label{tab:dataset}
\end{table}



 


\subsection{Analysis of Annotation Distributions}

\paragraph{Flagged Segments} The actual number of annotated segments is 10,896; but 2.3\% were excluded for flagged problems, and 0.13\% excluded for flagged difficulties. 

\paragraph{Readership Groups and Readability Levels} 
 Figure~\ref{fig:annotation-distribution} visualizes the annotation distributions across the four readership groups identified based on educated guesses and self-declared target readers. Full details are in Appendix~\ref{app:annotation-stats}. Children's texts dominate the easier levels ({\BlevAlif} to {\BlevHa}), while Classical texts dominate the harder levels ({\BlevSad} and {\BlevQaf}), as expected. The middle levels contain a mix of all groups. Interestingly, some Children texts include advanced materials, which may need revision, or can be arguably justified for educational purposes.

\paragraph{Readability Level Patterns} 
In terms of total counts, Figure~\ref{fig:annotation-distribution} exhibits a slightly skewed distribution, notably with lower counts for {\BlevTa} and higher counts for {\BlevNun}. This pattern could stem from the limited sample size or potential biases in text selections. Notably, the guidelines for {\BlevTa} feature specific uncommon linguistic elements like the dual command form, vocative, emotional vocabulary, and the Hamza interrogative particle.

\paragraph{Readability Level and Text Length} 

Figure~\ref{fig:annotation-level-avg} presents two charts comparing readability levels with segment lengths. The overall averages show a generally expected linear pattern from {\BlevAlif} to \BlevYa/\BlevKaf, continuing to {\BlevAyn} before dropping off, as higher-level texts, often poetry, are shorter than prose. The length distribution chart, in Figure~\ref{fig:annotation-level-avg}(right), highlights variability within each readability level, confirming that annotators did not strictly use segment lengths for readability level annotation.

\begin{table}
\centering
\tabcolsep2pt
\begin{small}
\begin{tabular}{|c|c|c|c|}
\hline
\textbf{Metric} &\textbf{CAMeLBERT} &\textbf{MARBERT} &\textbf{AraBERT} \\
\hline
\textbf{Accuracy @1} &\textbf{58\%} &56\% &57\% \\
\hline
\textbf{Accuracy @2 }&\textbf{73\%} &72\% &\textbf{73\%} \\
\hline
\textbf{Accuracy @3} &\textbf{83\%} &82\% &82\% \\
\hline
\textbf{CL Rank} &\textbf{2.23} &2.31 &2.24 \\
\hline
\textbf{CL Distance} &\textbf{1.06} &1.10 &1.07 \\
\hline
\textbf{QWK} &\textbf{84\%} &\textbf{84\%} &\textbf{84\%} \\
\hline
\end{tabular}
\end{small}
\caption{Results of automatic readability assessment comparing
CAMeLBERT-MIX, MARBERT, and AraBERTv2. CL Rank is the average rank of the correct label; CL Distance is the average distance from the correct label; and QWK is the Quadratic Weighted Kappa.}
\label{tab:ai-exp}
\end{table}

\subsection{Automatic Readability Assessment}
We train sentence-level classifiers by finetuning CAMeLBERT-MIX \cite{inoue-etal-2021-interplay}, MARBERT \cite{abdul-mageed-etal-2021-arbert} and AraBERTv2 \cite{antoun-etal-2020-arabert} to benchmark the baseline performance given the dataset.
We split the dataset into $90\%$ for training and $10\%$ for testing.
We finetune the models using the Transformers library \cite{Wolf:2019:huggingfaces} on a NVIDIA T4 GPU for three epochs with a learning rate of 5e-5, and a batch size of 16.
Table~\ref{tab:ai-exp} shows the results of finetuning the three models for readability prediction as a text classification task.
%
%
We report with the following metrics:
 \textbf{Accuracy@n} (correct label is within the top~$n$ predictions),
 \textbf{Average Rank of the Correct Label},
 \textbf{Average Distance from Correct Label}, and \textbf{Quadratic Weighted Kappa}.
The performance of the compared systems is generally similar. Their results are comparable with the IAA numbers, showing a robust Quadratic Weighted Kappa score of 84\%. We anticipate that performance will improve further with additional data.

}

\subsection{Error Analysis}
\label{sec:error-analysis}
To assess the errors in our best-performing model, we analyzed error patterns in the inter-annotator portion of the development (DEV) set.
Each sentence in this subset had five human annotations, which we compared to the model’s prediction.

We grouped sentences by the level of annotator agreement, from full agreement (5 out of 5 annotators) down to minimal agreement (1 out of 5).
Full 5-way agreement accounts for 25\% of the data.
With each reduction in agreement -- to 4, 3, 2, and finally 1 annotator -- the cumulative coverage increases to 50\%, 61\%, 72\%, and 87\%, respectively.
In other words, in 87\% of the cases, the model prediction can be meaningfully compared to at least some level of human consensus.

The remaining 13\% fall outside this range.
In 1\% of these, the model’s prediction was within the span of human annotations but did not exactly match any of them.
In 3\%, the prediction was above the maximum annotation, and in 9\%, it was below the minimum.
We manually reviewed these out-of-range cases and found that the annotators were generally correct.
We speculate that the model’s errors arise from limited training data, lack of contextual understanding, or insufficient modeling of linguistic features.
For example, the obscure word \<عصامة> \textit{{\AYN}SAm{\TAMAR}} ‘tightly wound head dress’ may be misinterpreted as the feminine form of the proper name \<عصام> \textit{{\AYN}SAm} ‘Esam’, much like connecting \<كريم> 
\textit{krym} `Kareem' with \<كريمة> \textit{krym{\TAMAR}} `Kareema'. However, \<عصامة> \textit{{\AYN}SAm{\TAMAR}} is not a plausible proper name. This remains speculative, as our model is not inherently interpretable.

\section{Conclusions and Future Work}

This paper presented the \textbf{Balanced Arabic Readability Evaluation Corpus (BAREC)}, a large-scale, finely annotated dataset for assessing Arabic text readability across 19 levels. With over 69K sentences and 1 million words, it is the largest Arabic corpus for readability assessment, covering diverse genres, topics, and audiences, to our knowledge. High inter-annotator agreement ensures reliable annotations. Through benchmarking various readability assessment techniques, we highlighted both the challenges and opportunities in Arabic readability modeling, demonstrating promising performance across different methods.

Looking ahead, we plan to expand the corpus, enhancing its size and diversity to cover additional genres and topics. We also aim to add annotations related to vocabulary leveling and syntactic treebanks to study less-explored genres in syntax. Future work will include analyzing readability differences across genres and topics. Additionally, the tools we have developed will be integrated into a system to help children's story writers target specific reading levels.

The {\barec} dataset, its annotation guidelines, and benchmark results, will be made publicly available to support future research and educational applications in Arabic readability assessment.

\newpage

\section*{Acknowledgments}
The {\barec} project is supported by the Abu Dhabi Arabic Language Centre (ALC) / Department of Culture and Tourism, UAE. 
We acknowledge the support of the High Performance Computing Center at New York University Abu Dhabi.
We are deeply grateful to our outstanding annotation team:  Mirvat Dawi, Reem Faraj, Rita Raad, Sawsan Tannir, and Adel Wizani, Samar Zeino, and Zeina Zeino. 
Special thanks go to Abdallah Abushmaes, Karin Aghadjanian, and Omar Al Ayyoubi of the ALC for their continued support.
We would also like to thank the Zayed University ZAI Arabic Language Research Center team, in particular Hamda Al-Hadhrami,  Maha Fatha, and Metha Talhak, for their valuable contributions to typing materials for the project.  We also acknowledge Ali Gomaa and his team for their additional support in this area.
Finally, we thank our colleagues at the New York University Abu Dhabi Computational Approaches to Modeling Language (CAMeL) Lab,  Muhammed Abu Odeh,  Bashar Alhafni, Ossama Obeid, and Mostafa Saeed, as well as Nour Rabih (Mohamed bin Zayed University of Artificial Intelligence)  for their helpful conversations and feedback.

\section*{Limitations}
One notable limitation is the inherent subjectivity associated with readability assessment, which may introduce variability in annotation decisions despite our best efforts to maintain consistency. Additionally, the current version of the corpus may not fully capture the diverse linguistic landscape of the Arab world. Finally, while our methodology strives for inclusivity, there may be biases or gaps in the corpus due to factors such as selection bias in the source materials or limitations in the annotation process. We acknowledge that readability measures   can be used with malicious intent to profile people; this is not our intention, and we discourage it.

\section*{Ethics Statement}

All data used in the corpus curation process are sourced responsibly and legally.
The annotation process is conducted with transparency and fairness, with multiple annotators involved to mitigate biases and ensure reliability. All annotators are paid fair wages for their contribution.  The corpus and associated guidelines are made openly accessible to promote transparency, reproducibility, and collaboration in Arabic language research.


\bibliography{custom,camel-bib-v3,anthology}

\onecolumn
\appendix
\section{{\barec} Annotation Guidelines Cheat Sheet and Examples}
\label{app:guidelines}
\subsection{Arabic Original}
\begin{center}
  \includegraphics[width=\textwidth]{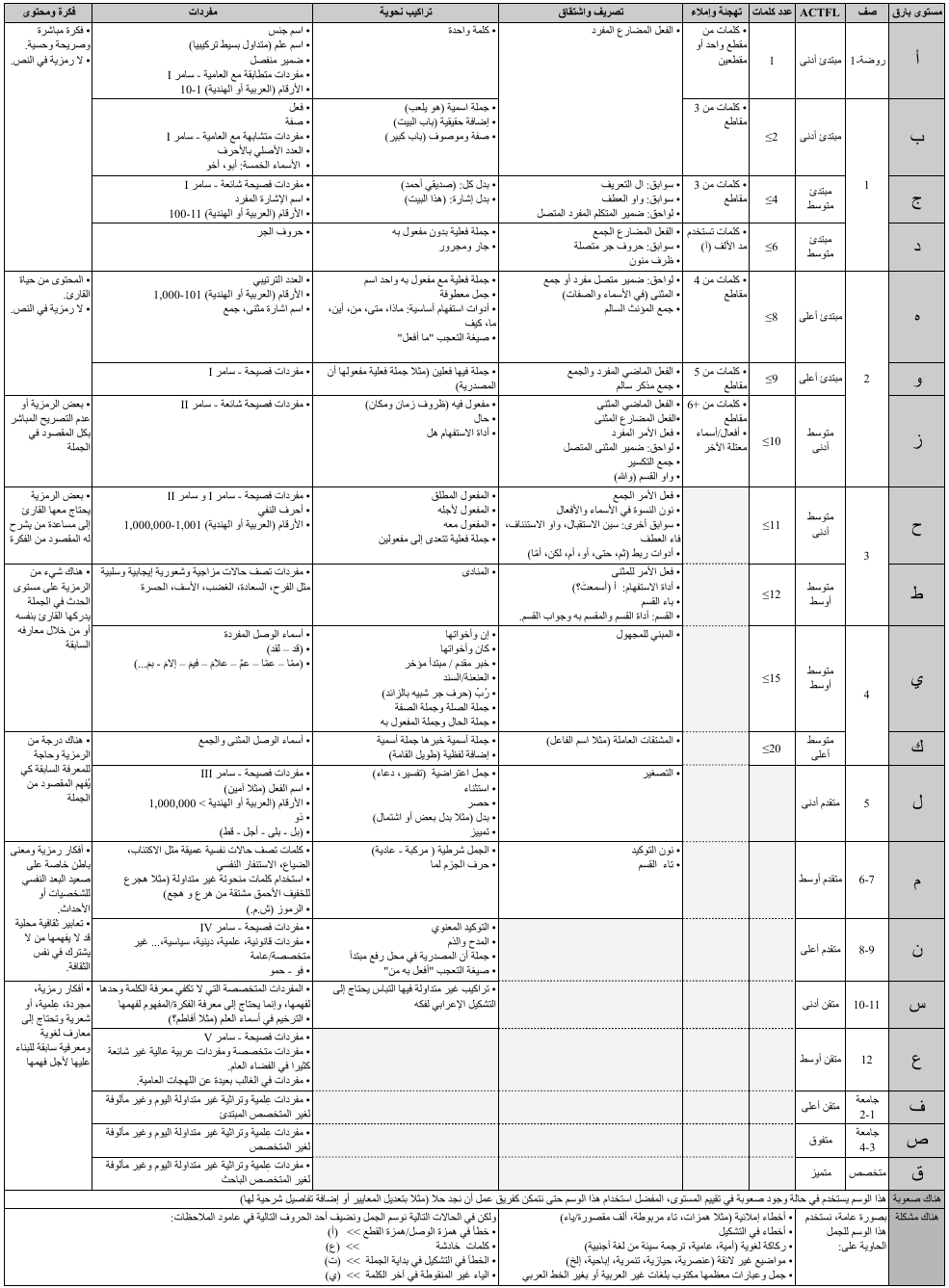}
  \end{center} 
\newpage 
\subsection{English Translation}

\begin{center}
  \includegraphics[width=\textwidth]{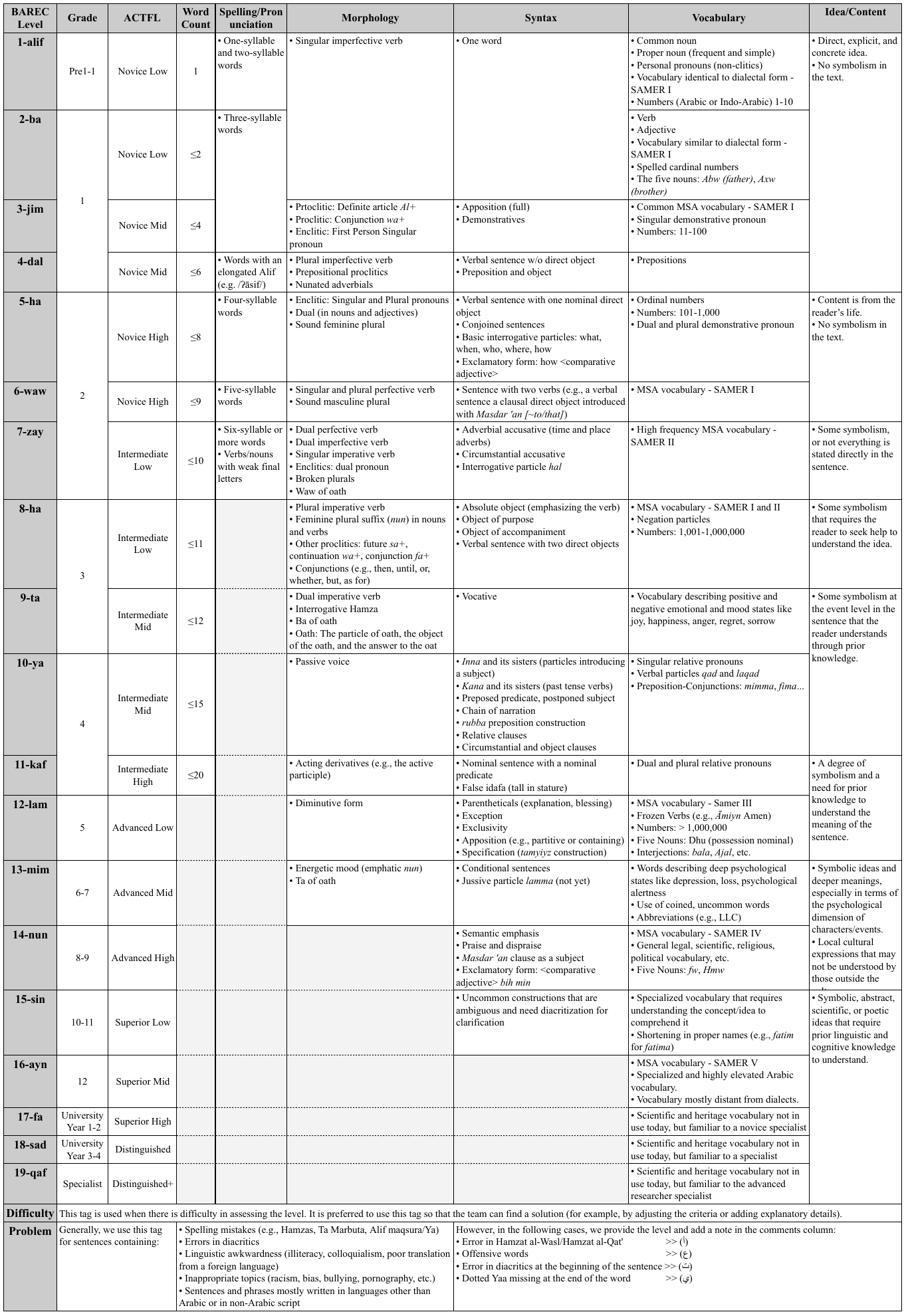}
\end{center} 

\newpage 
\subsection{Annotation Examples}
\label{fullexample}
Representative examples of the 19 {\barec} readability levels, with English translations, and readability level reasoning. Underlining is used to highlight the main keys that determined the level. 

  \includegraphics[width=\textwidth]{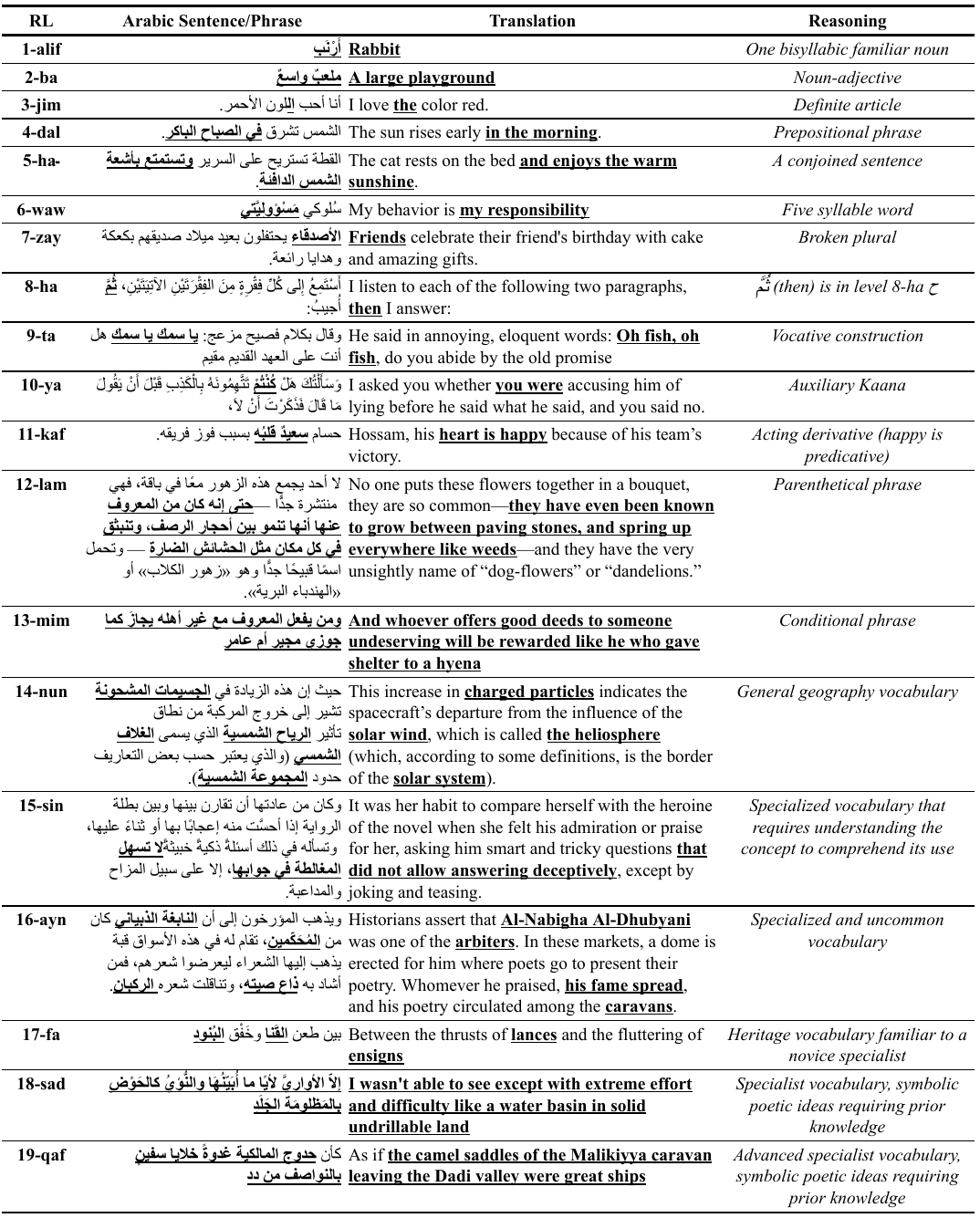}
\newpage 

\section{{\barec} Corpus Splits}
\label{app:splits}
\subsection{Sentence-level splits across readability levels}
\begin{center}
  \includegraphics[]{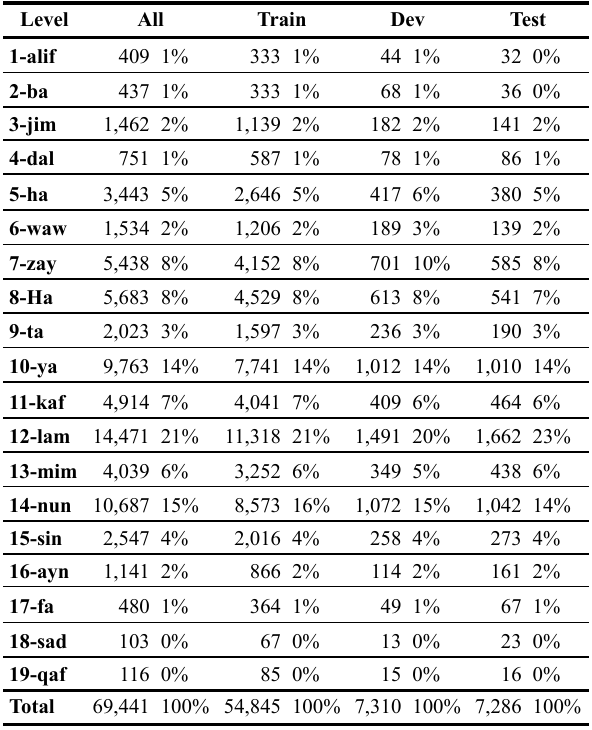}
  \end{center} 
\newpage 
\subsection{Sentence-level splits across domains and readership groups}
\begin{center}
  \includegraphics[]{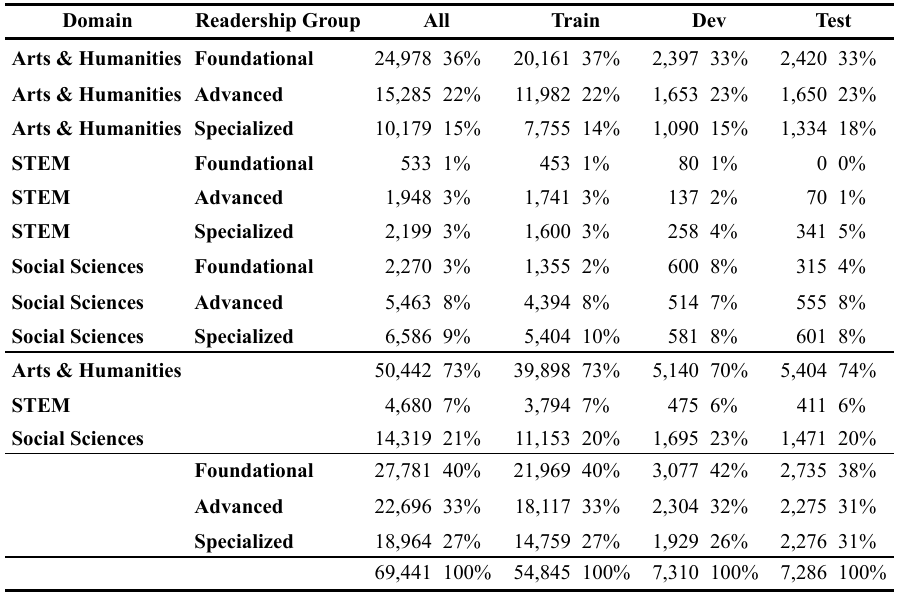}
\end{center}

\twocolumn
\section{{\barec} Corpus Details}
\label{app:full-data}

\subsection{Resources}
 
We present the corpus sources in groups of their general intended purpose.

\subsubsection{Education}

\paragraph{Emarati Curriculum} The first five units of the UAE curriculum textbooks for the 12 grades in three subjects: Arabic language, social studies, Islamic studies \cite{Khalil:2018:leveled}.

\paragraph{ArabicMMLU} 6,205 question and answer pairs from the ArabicMMLU benchmark dataset \cite{koto-etal-2024-arabicmmlu}.

\paragraph{Zayed Arabic-English Bilingual Undergraduate Corpus (ZAEBUC)}
100 student-written articles from the Zayed University Arabic-English Bilingual Undergraduate Corpus \cite{habash-palfreyman-2022-zaebuc}.

\paragraph{Arabic Learner Corpus (ALC)}
16 L2 articles from the Arabic Learner Corpus \citep{phdthesis}.

\paragraph{Basic Travel Expressions Corpus (BTEC)}
20 documents from the MSA translation of the Basic Traveling Expression Corpus \cite{eck-hori-2005-overview,takezawa-etal-2007-multilingual,bouamor-etal-2018-madar}.

\paragraph{Collection of Children poems} Example of the included poems: My language sings (\<لغتي تغني>), and Poetry and news (\<أشعار وأخبار>) \cite{kashkol, poetry-and-news}.

\paragraph{ChatGPT} To add more children's materials, we ask Chatgpt to generate 200 sentences ranging from 2 to 4 words per sentence, 150 sentences ranging from 5 to 7 words per sentence and 100 sentences ranging from 8 to 10 words per sentence.\footnote{\url{https://chatgpt.com/}} Not all sentences generated by ChatGPT were correct. We discarded some sentences that were flagged by the annotators. Table~\ref{tab:chatgpt} shows the prompts and the percentage of discarded sentences for each prompt.

\begin{table*}[t!]
\centering
  \includegraphics[width=1.8\columnwidth]{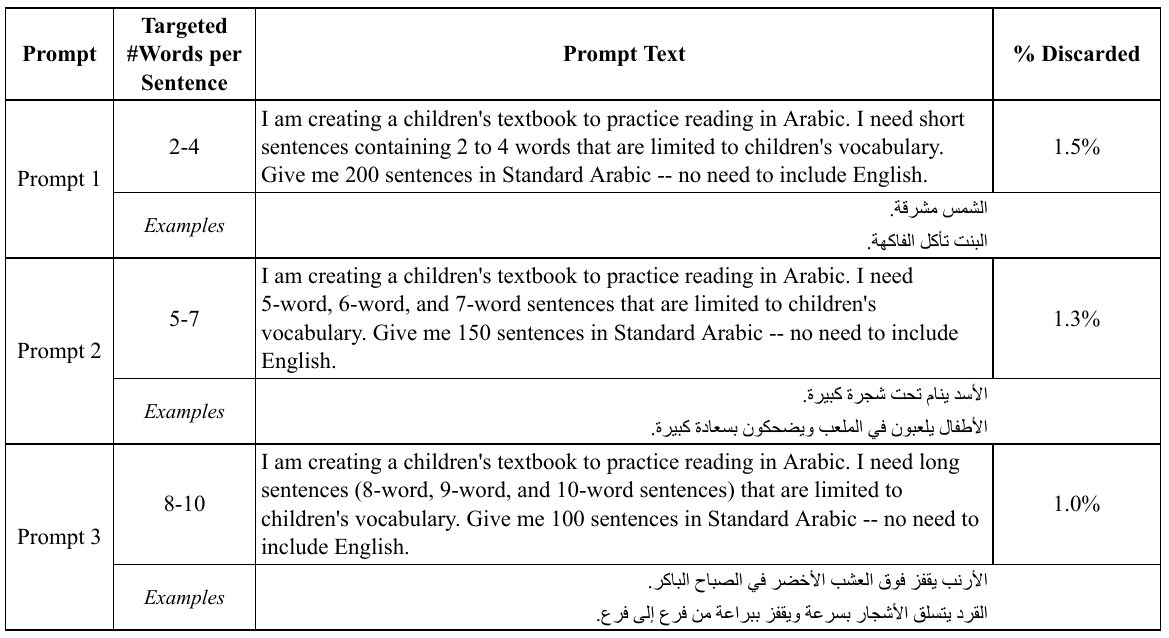}
    \caption{ChatGPT Prompts. \% Discarded is the percentage of discarded sentences due to grammatical errors.}
\label{tab:chatgpt}
\end{table*}

\subsubsection{Literature}

\paragraph{Hindawi} A subset of 264 books extracted from the Hindawi Foundation website across different different genres.\footnote{\url{https://www.hindawi.org/books/categories/}}

\paragraph{Kalima}
The first 500 words of 62 books from Kalima project.\footnote{\url{https://alc.ae/publications/kalima/}}

\paragraph{Green Library}
58 manually typed books from the Green Library.\footnote{\url{https://archive.org/details/201409_201409}}

\paragraph{Arabian Nights} The openings and endings of the opening narrative and the first eight nights from the Arabian Nights \cite{ArabianNights}. We extracted the text from an online forum.\footnote{\url{http://al-nada.eb2a.com/1000lela\&lela/}}

\paragraph{Hayy ibn Yaqdhan}  A subset of the philosophical novel and allegorical tale written by Ibn Tufail \cite{tufail:hayy}.
We extracted the text from the Hindawi Foundation website.\footnote{\url{https://www.hindawi.org/books/90463596/}}

\paragraph{Sara} The first 1000 words of {\it Sara}, a novel by Al-Akkad first published in 1938 \cite{akkad:sarah}. We extracted the text from the Hindawi Foundation website.\footnote{\url{https://www.hindawi.org/books/72707304/}}

\paragraph{The Suspended Odes (Odes)}  The ten most celebrated poems from Pre-Islamic Arabia (\<المعلقات> Mu’allaqat).
All texts were extracted from Wikipedia.\footnote{\url{https://ar.wikipedia.org/wiki/}\<المعلقات>}

\subsubsection{Media}

\paragraph{Majed}
10 manually typed editions of Majed magazine for children from 1983 to 2019.\footnote{\url{https://archive.org/details/majid_magazine}}

\paragraph{ReadMe++}
The Arabic split of the ReadMe++ dataset \cite{naous-etal-2024-readme}.

\paragraph{Spacetoon Songs}
The opening songs of 53 animated children series from Spacetoon channel.

\paragraph{Subtitles} A subset of the Arabic side of the OpenSubtitles 
dataset \cite{Lison:2016:opensubtitles2016}.

\paragraph{WikiNews} 62 Arabic WikiNews articles covering politics, economics,
health, science and technology, sports, arts, and culture \cite{Abdelali:2016:farasa}.

\subsubsection{References} 
\paragraph{Wikipedia} A subset of 168 Arabic wikipedia articles covering Culture, Figures, Geography, History, Mathematics, Sciences, Society, Philosophy, Religions and Technologies.\footnote{\url{https://ar.wikipedia.org/}}

\paragraph{Constitutions}
The first 2000 words of the Arabic constitutions from 16 Arabic speaking countries, collected from MCWC dataset \cite{el-haj-ezzini-2024-multilingual}.

\paragraph{UN} The Arabic translation of the Universal Declaration of Human Rights.\footnote{\url{https://www.un.org/ar/about-us/universal-declaration-of-human-rights}}

\subsubsection{Religion}

\paragraph{Old Testament} The first 20 chapters of the Book of  Genesis \cite{arabicOldTestament}.\footnote{\url{https://www.arabicbible.com/}\label{biblefoot}}

\paragraph{New Testament} The first 16 chapters of the Book of Matthew \cite{arabicNewTestament}.$^{\ref{biblefoot}}$ 

\paragraph{Quran} The first three Surahs and the last 14 Surahs from the Holy Quran. We selected the text from the Quran Corpus Project \cite{Dukes:2013:supervised}.\footnote{\url{https://corpus.quran.com/}}

\paragraph{Hadith} The first 75 Hadiths from Sahih Bukhari \cite{bukhari}.  We selected the text from the LK Hadith Corpus\footnote{\url{https://github.com/ShathaTm/LK-Hadith-Corpus}} \cite{Altammami:2019:Arabic}.

Some datasets are chosen because they already have annotations available for other tasks.
For example, dependency treebank annotations exist for \textbf{Odes}, \textbf{Quran}, \textbf{Hadith}, \textbf{1001}, \textbf{Hayy}, \textbf{OT}, \textbf{NT}, \textbf{Sara},\textbf{WikiNews}, \textbf{ALC}, \textbf{BTEC}, and \textbf{ZAEBUC} \cite{habash-etal-2022-camel}.

\subsection{Domains}
\label{app:domains}
\paragraph{Arts \& Humanities}
The Arts and Humanities domain comprised the following subdomains.
\begin{itemize}
\item  \textit{Literature and Fiction:} Encompasses novels, short stories, poetry, and other creative writing forms that emphasize narrative and artistic expression.	
\item  \textit{Religion and Philosophy: }Contains religious texts, philosophical works, and related writings that explore spiritual beliefs, ethics, and metaphysical ideas.	
\item  \textit{Education and Academic Texts (on Arts and Humanities):} Includes textbooks, scholarly articles, and educational materials that are often structured for learning and academic purposes.	
\item  \textit{General Knowledge and Encyclopedic Content (on Arts and Humanities): } Covers reference materials such as encyclopedias, almanacs, and general knowledge articles that provide broad information on various topics.	
\item  \textit{News and Current Affairs (on Arts and Humanities):} Includes newspapers, magazines, and online news sources that report on current events and issues affecting society.	
\end{itemize}

\paragraph{Social Sciences}
The Social Sciences domain comprised the following subdomains.
\begin{itemize}
    \item  \textit{Business and Law:} Encompasses legal texts, business strategies, financial reports, and corporate documentation relevant to professional and legal contexts.	
    \item  \textit{Social Sciences and Humanities:} Covers disciplines like sociology, anthropology, history, and cultural studies, which explore human society and culture.	
    \item  \textit{Education and Academic Texts (on Social Sciences):} Includes textbooks, scholarly articles, and educational materials that are often structured for learning and academic purposes.	
    \item  \textit{General Knowledge and Encyclopedic Content (on Social Sciences):} Covers reference materials such as encyclopedias, almanacs, and general knowledge articles that provide broad information on various topics.	
    \item  \textit{News and Current Affairs (on Social Sciences):} Includes newspapers, magazines, and online news sources that report on current events and issues affecting society.	
\end{itemize}

\paragraph{STEM}
 	
 The Science, Technology, Engineering and Mathematics domain comprised the following subdomains.

\begin{itemize}
    \item \textit{Science and Technology:} Includes scientific research papers, technology articles, and technical manuals that focus on advancements and knowledge in science and tech fields.	
    \item \textit{Education and Academic Texts (on STEM):} Includes textbooks, scholarly articles, and educational materials that are often structured for learning and academic purposes.	
    \item \textit{General Knowledge and Encyclopedic Content (on STEM):} Covers reference materials such as encyclopedias, almanacs, and general knowledge articles that provide broad information on various topics.	
    \item \textit{News and Current Affairs (on STEM):} Includes newspapers, magazines, and online news sources that report on current events and issues affecting society.	
\end{itemize}

\subsection{Readership Groups}
\label{app:reader}
\paragraph{Foundational}
This level includes learners, typically up to 4th grade or age 10, who are building basic literacy skills, such as decoding words and understanding simple sentences.	

\paragraph{Advanced}
Refers to individuals with average adult reading abilities, capable of understanding a variety of texts with moderate complexity, handling everyday reading tasks with ease.	

\paragraph{Specialized}
Represents readers with advanced skills, typically starting in 9th grade or above in specialized topics, who can comprehend and engage with complex, domain-specific texts in specialized fields.

\onecolumn

\begin{table*}[ht!]
\begin{center}
    \includegraphics[scale=0.96]{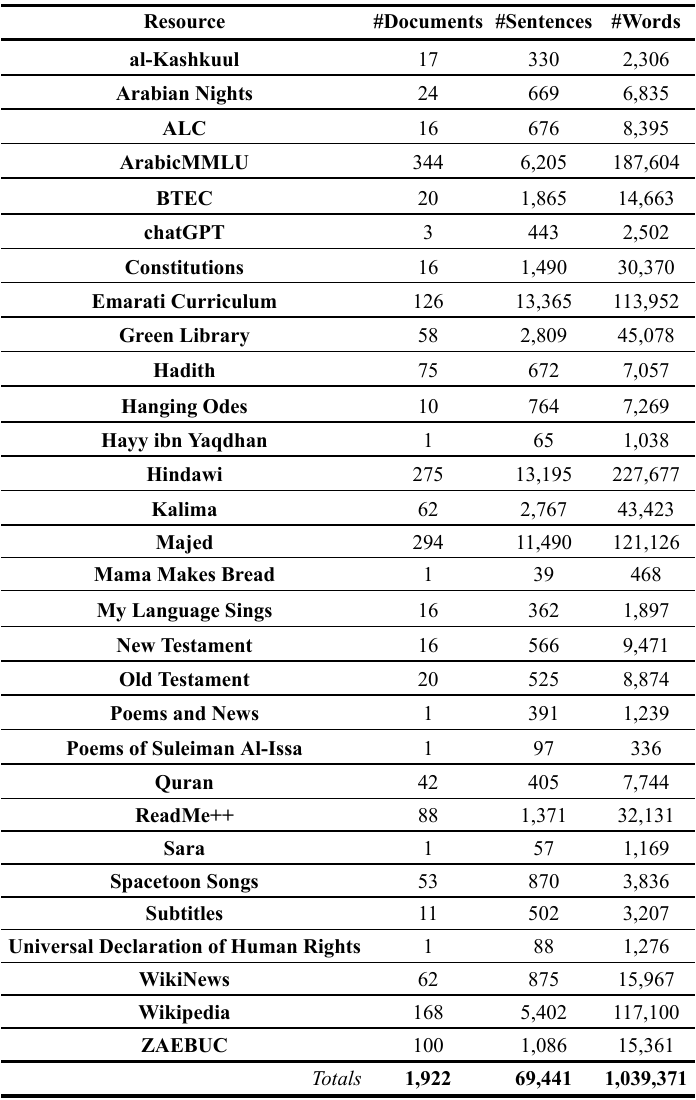}
    \caption{{\barec} Corpus Details: the texts used to build the dataset, and the number of documents, sentences, and words extracted from each text.}
\end{center}
  
\label{tab:dataset-details}
\end{table*}




\newpage
\onecolumn
\section{Additional Results}
\subsection{Confusion Matrix}
\label{app:conf-matrix}

\begin{figure*}[h!]
\centering
 \includegraphics[width=0.7\columnwidth]{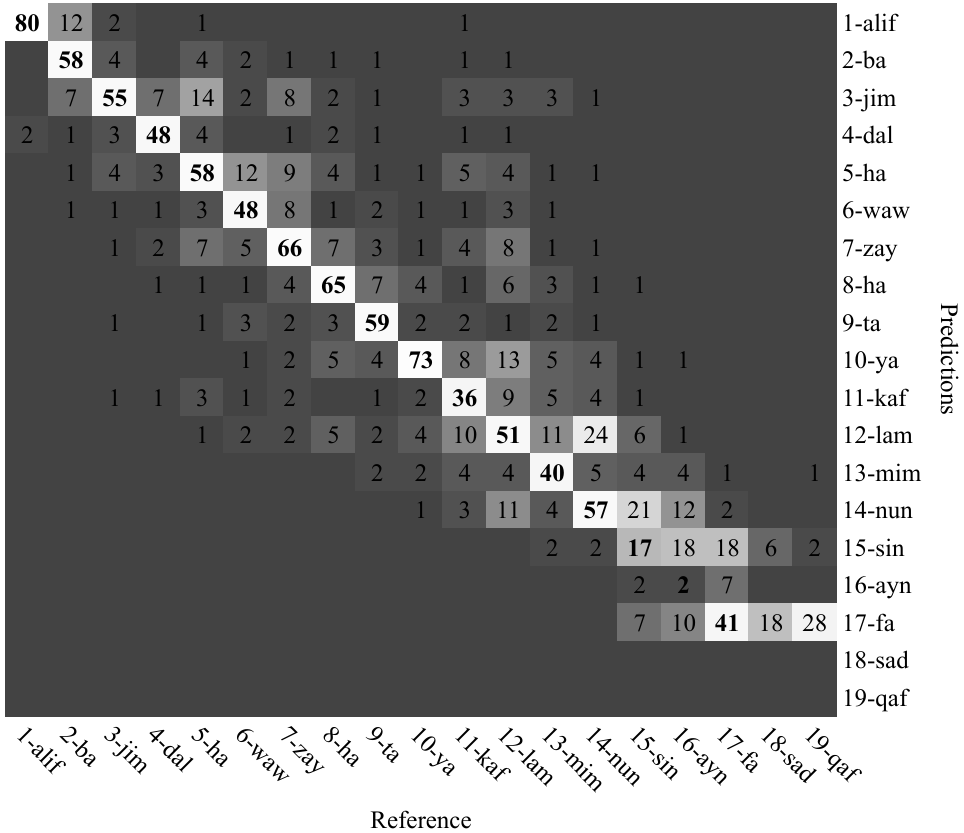}
    \caption{Confusion matrix of F-score across the different readability levels for the best model from stage 1.}
\label{fig:conf-matrix}
\end{figure*}

Figure~\ref{fig:conf-matrix} shows the confusion matrix for the best-performing model from Stage 1: the AraBERTv2 model trained on D3Tok sentences with Cross-Entropy (CE) loss. The matrix uses F-scores to account for the unbalanced distribution of readability levels. The strong diagonal indicates a high rate of exact matches between predicted and gold labels. However, the model exhibits more disagreement at the higher, more difficult levels— likely due to the scarcity of training examples in those levels. Additionally, the model shows a tendency to under-estimate readability levels, favoring lower labels. This aligns with the patterns observed in the error analysis discussed in Section~\ref{sec:error-analysis}.

\newpage
\subsection{All Loss Functions}
\label{app:loss}

\begin{table*}[h!]
\centering
\begin{tabular}{llcccc}
\toprule
\textbf{Input} & \textbf{Model} & \textbf{Acc$^{19}$} & \textbf{$\pm$1 Acc$^{19}$} & \textbf{Dist} & \textbf{QWK} \\
\midrule
\textbf{Word}   & SVM            & 36.2\% & 47.9\% & 2.03 & 53.4\% \\
\textbf{D3Tok}  & SVM            & 37.2\% & 49.3\% & 1.92 & 56.5\% \\
\textbf{Word}   & DecisionTree   & 27.2\% & 41.2\% & 2.50 & 44.2\% \\
\textbf{D3Tok}  & DecisionTree   & 29.9\% & 44.2\% & 2.33 & 52.5\% \\
\midrule
\textbf{D3Tok} & AraBERTv2 &&&& \\
       & +CE            & \textbf{56.6\%} & 69.9\% & 1.14 & 80.0\% \\
       & +EMD           & 55.3\% & 70.3\% & \textbf{1.11} & 81.2\% \\
       & +OLL2          & 35.2\% & 70.3\% & 1.25 & 82.0\% \\
       & +OLL15         & 47.3\% & 71.1\% & 1.13 & 82.8\% \\
       & +OLL1          & 50.8\% & 71.5\% & 1.12 & 81.7\% \\
       & +OLL05         & 53.1\% & 68.8\% & 1.18 & 79.7\% \\
       & +SOFT2         & 55.8\% & 69.8\% & 1.15 & 80.0\% \\
       & +SOFT3         & 56.4\% & 69.9\% & 1.14 & 80.1\% \\
       & +SOFT4         & 56.4\% & 69.9\% & 1.15 & 79.6\% \\
       & +SOFT5         & 56.2\% & 69.5\% & 1.17 & 79.3\% \\
       & +Reg     & 43.1\% & \textbf{73.1\%} & 1.13 & \textbf{84.0\%} \\
\bottomrule
\end{tabular}
\caption{Loss functions comparisons on {\barec} Dev set. For SVM and Decision Tree classifiers, we used count vectorizer.}
\label{tab:loss-functions-full}
\end{table*}




\subsection{Impact of Training Granularity on Readability Level Prediction}
\label{app:gran}

To analyze the effect of training granularity on readability level prediction, we compare two approaches: (1) training on all 19 levels and then mapping predictions to lower levels (7, 5, or 3), and (2) training directly on the target granularity.

Table \ref{tab:gran} presents the results of this comparison.
Overall, training on 19 levels and then mapping achieves slightly better performance across for 5-level and 3-level granularities compared to direct training. Moreover, the performance gap between the two approaches widens as the target granularity becomes coarser, suggesting that finer-grained supervision during training provides more informative learning signals, which translate into improved generalization when predictions are mapped into broader scales.

\begin{table*}[h!]
\centering
\begin{tabular}{ccllcccc}
\toprule
\textbf{Train Gran} & \textbf{Dev Gran} & \textbf{Input} & \textbf{Model} & \textbf{Acc} & $\pm$\textbf{1 Acc} & \textbf{Dist} & \textbf{QWK} \\
\midrule
19 & 7 & D3Tok & CE & \textbf{65.9\%} & 88.9\% & 0.51 & 79.9\% \\
7  & 7 & D3Tok & CE & 65.2\% & \textbf{89.5\%} & \textbf{0.50} & \textbf{81.0\%} \\
\midrule
19 & 5 & D3Tok & CE & \textbf{70.3\%} & 93.5\% & \textbf{0.37} & \textbf{78.3\%} \\
5  & 5 & D3Tok & CE & 67.8\% & \textbf{93.7\%} & 0.39 & 77.3\% \\
\midrule
19 & 3 & D3Tok & CE & \textbf{76.5\%} & \textbf{97.6\%} & \textbf{0.26} & \textbf{74.7\%} \\
3  & 3 & D3Tok & CE & 74.4\% & 96.9\% & 0.29 & 74.0\% \\
\bottomrule
\end{tabular}
\caption{Comparison between training on 19 levels then mapping to the target granularity vs. training directly on the target granularity.}
\label{tab:gran}
\end{table*}

\newpage
\onecolumn
\subsection{Ensembles \& Oracles}
\label{app:ensemble}

\begin{table*}[h!]
\centering
\begin{tabular}{cccccc|cccc}
\toprule
\textbf{CE} & \textbf{CE} & \textbf{CE} & \textbf{CE} & \textbf{EMD} & \textbf{Reg} & \multicolumn{4}{c}{\textbf{Metrics}} \\
 \cmidrule(lr){7-10}
\textbf{Word} & \textbf{Lex} & \textbf{D3Tok} & \textbf{D3Lex} & \textbf{D3Tok} & \textbf{D3Tok} 
& \textbf{Acc$^{19}$} & \textbf{$\pm$1 Acc$^{19}$} & \textbf{Dist} & \textbf{QWK} \\
\midrule
\checkmark &           &           &           &           &           & 51.6\% & 65.9\% & 1.32 & 76.3\% \\
          & \checkmark &           &           &           &           & 50.1\% & 65.4\% & 1.29 & 77.7\% \\
          &           & \checkmark &           &           &           & \textbf{56.6\%} & 69.9\% & 1.14 & 80.0\% \\
          &           &           & \checkmark &           &           & 53.2\% & 67.1\% & 1.24 & 78.6\% \\
          &           &           &           & \checkmark &           & 55.3\% & 70.3\% & \textbf{1.11} & 81.2\% \\
          &           &           &           &           & \checkmark & 43.1\% & \textbf{73.1\%} & 1.13 & \textbf{84.0\%} \\
\midrule
\multicolumn{6}{l|}{Average}        & 46.9\% & 72.5\% & \textbf{1.11} & 83.4\% \\
\multicolumn{6}{l|}{Most Common}   & 56.3\% & 70.0\% & 1.13 & 80.4\% \\
\midrule
\multicolumn{6}{l|}{Oracle Combinations} & & & & \\
\checkmark & \checkmark &           &           &           &           & 62.4\% & 76.6\% & 0.88 & 88.4\% \\
\checkmark &           & \checkmark &           &           &           & 63.5\% & 76.7\% & 0.89 & 87.7\% \\
\checkmark &           &           & \checkmark &           &           & 63.2\% & 76.6\% & 0.88 & 88.2\% \\
\checkmark &           &           &           & \checkmark &           & 63.3\% & 77.9\% & 0.83 & 89.2\% \\
\checkmark &           &           &           &           & \checkmark & 62.0\% & 80.7\% & 0.77 & 90.8\% \\
\checkmark & \checkmark & \checkmark & \checkmark &           &           & 69.5\% & 82.3\% & 0.67 & 91.4\% \\
\checkmark & \checkmark & \checkmark & \checkmark & \checkmark &           & 72.0\% & 84.5\% & 0.59 & 92.6\% \\
\checkmark & \checkmark & \checkmark & \checkmark &           & \checkmark & 73.6\% & 86.6\% & 0.53 & 93.4\% \\
\checkmark & \checkmark & \checkmark & \checkmark & \checkmark & \checkmark & 75.2\% & 87.4\% & 0.50 & 93.8\% \\
\bottomrule
\end{tabular}
\caption{Comparison between individual models, ensembles and oracles on {\barec} Dev set.}
\label{tab:oracle-ensemble}
\end{table*}


    





\hide{
\section{Detailed Annotation Stats}
\label{app:annotation-stats}
\centering
\begin{table*}[h!]
\centering
  \includegraphics[scale=0.7]{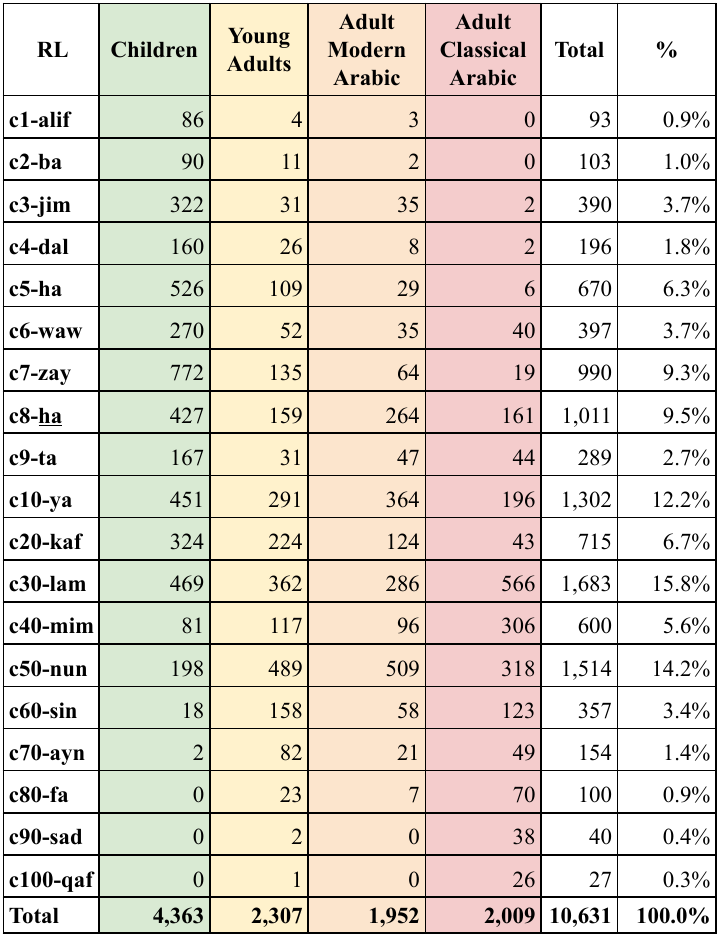}
    \caption{Detailed Annotation Statistics across Readability Levels and Reading Groups.}
\label{tab:annotation-stats}
\end{table*}
}

\end{document}